\documentclass[10pt,twocolumn,letterpaper]{article}

\usepackage{cvpr} 
\usepackage{times}
\usepackage{epsfig}
\usepackage{graphicx}
\usepackage{amsmath}
\usepackage{amssymb}
\usepackage{booktabs}
\usepackage{multirow}
\usepackage{array}
\usepackage{ mathrsfs }
\usepackage{float}
\usepackage{subcaption}
\usepackage{xcolor}
\usepackage{siunitx}
\usepackage[export]{adjustbox}% To produce the REVIEW version
\usepackage{graphicx}
\usepackage{amsmath}
\usepackage{amssymb}
\usepackage{booktabs}

\usepackage[pagebackref,breaklinks,colorlinks]{hyperref}
\usepackage[symbol,hang,flushmargin]{footmisc} 

\usepackage[capitalize]{cleveref}
\crefname{section}{Sec.}{Secs.}
\Crefname{section}{Section}{Sections}
\Crefname{table}{Table}{Tables}
\crefname{table}{Tab.}{Tabs.}

\newcommand{\klens}{K\textbar Lens GmbH, Germany}
\newcommand{\uni}{Saarland Informatics Campus, Germany}
\newcommand{\mpi}{MPI Informatics, Germany}
\newcommand{\dfki}{German Research Center for Artificial Intelligence (DFKI), Germany}

\makeatletter
\def\thanks#1{\protected@xdef\@thanks{\@thanks
		\protect\footnotetext{#1}}}
\def\cvprPaperID{21} % *** Enter the CVPR Paper ID here
\def\confName{CVPR}
\def\confYear{2023}

\usepackage{cuted}
\usepackage{capt-of}
\begin{document}

\title{ High-Resolution Synthetic RGB-D Datasets for Monocular Depth Estimation}

	\author{Aakash Rajpal$^{1,2}$$^{,}$, Noshaba Cheema$^{2,3,4}$, Klaus Illgner-Fehns$^{1}$, \\ Philipp Slusallek$^{2,4}$, and Sunil Jaiswal$^{1,*}$
		\vspace{0.5cm} \\
		$^{1}$ \klens, $^{2}$ \uni , \\ $^{3}$ \mpi, 
		  $^{4}$ \dfki   \\
		\thanks{		\hspace{-0.6cm} This work was partially funded by the German Ministry for Education and Research (BMBF) under the grant PLIMASC.}
	}
	\maketitle
	\footnotetext[1]{Corresponding author: {\tt\small sunil.jaiswal@k$-$lens.de}}

%%%%%%%%% ABSTRACT
\vspace{-1 cm}
\begin{abstract}
   Accurate depth maps are essential in various applications, such as autonomous driving, scene reconstruction, point-cloud creation, etc. However, monocular-depth estimation (MDE) algorithms often fail to provide enough texture \& sharpness, and also are inconsistent for homogeneous scenes. These algorithms mostly use CNN or vision transformer-based architectures requiring large datasets for supervised training. But, MDE algorithms trained on available depth datasets do not generalize well and hence fail to perform accurately in diverse real-world scenes. Moreover, the ground-truth depth maps are either lower resolution or sparse leading to relatively inconsistent depth maps. In general, acquiring a high-resolution ground truth dataset with pixel-level precision for accurate depth prediction is an expensive, and time-consuming challenge. 

In this paper, we generate a high-resolution synthetic depth dataset (HRSD) of dimension $1920 \times 1080$ from Grand Theft Auto (GTA-V), which contains 100,000 color images and corresponding dense ground truth depth maps. The generated datasets are diverse and have scenes from indoors to outdoors, from homogeneous surfaces to textures. For experiments and analysis, we train the DPT algorithm, a state-of-the-art transformer-based MDE algorithm on the proposed synthetic dataset, which significantly increases the accuracy of depth maps on different scenes by 9\%. Since the synthetic datasets are of higher resolution, we propose adding a feature extraction module in the transformer's encoder and incorporating an attention-based loss, further improving the accuracy by 15 \%. 
\end{abstract}
\vspace{-.5 cm}
%%%%%%%%% BODY TEXT
\section{Introduction}

\begin{figure}[h!]
\begin{subfigure}{.49\textwidth}
  \includegraphics[width=0.48\linewidth, height=25mm]{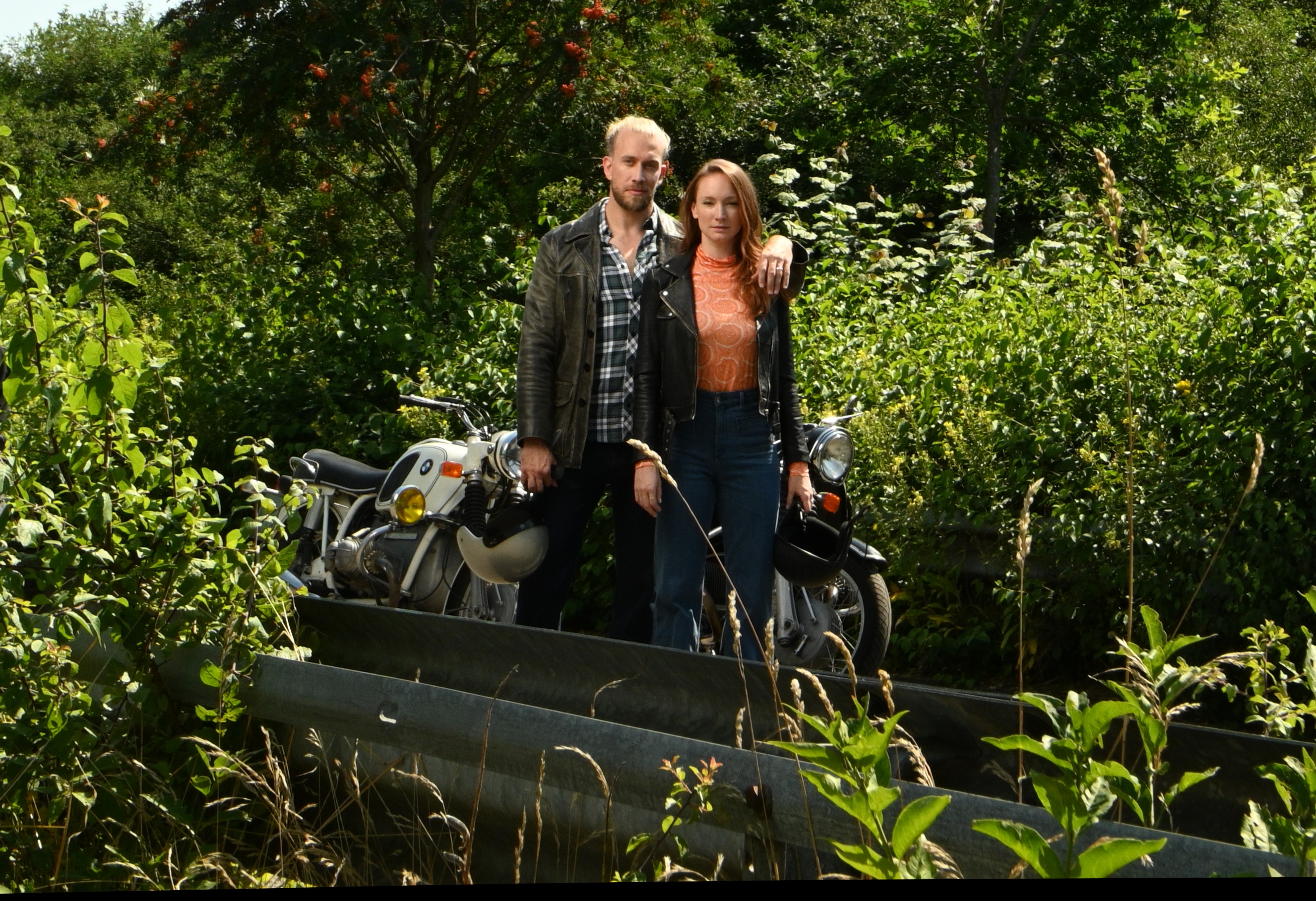}  
   \includegraphics[width=0.48\linewidth, height=25mm]{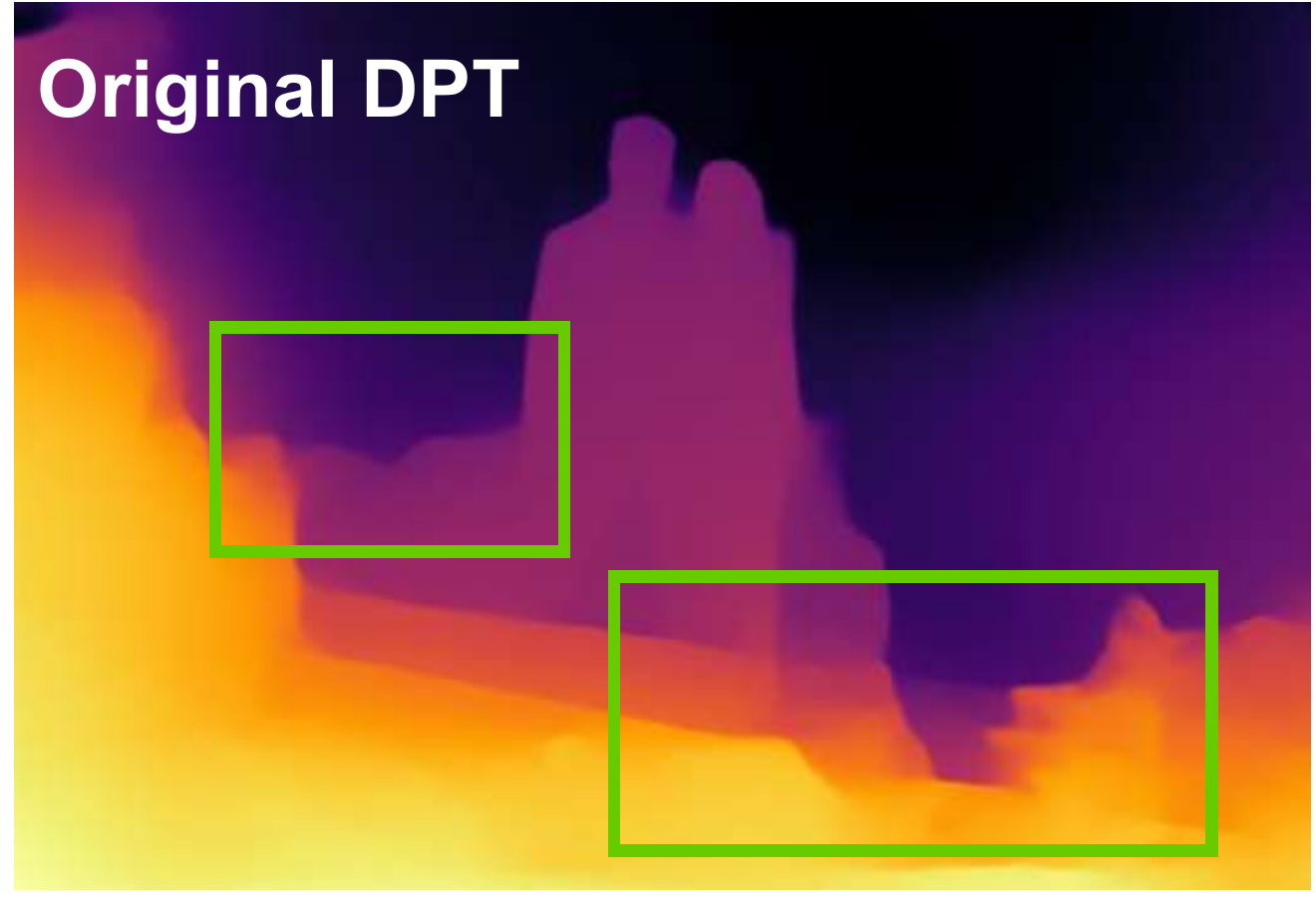} 
\end{subfigure}
\hfill
\begin{subfigure}{.49\textwidth}
  \includegraphics[width=0.48\linewidth, height=25mm]{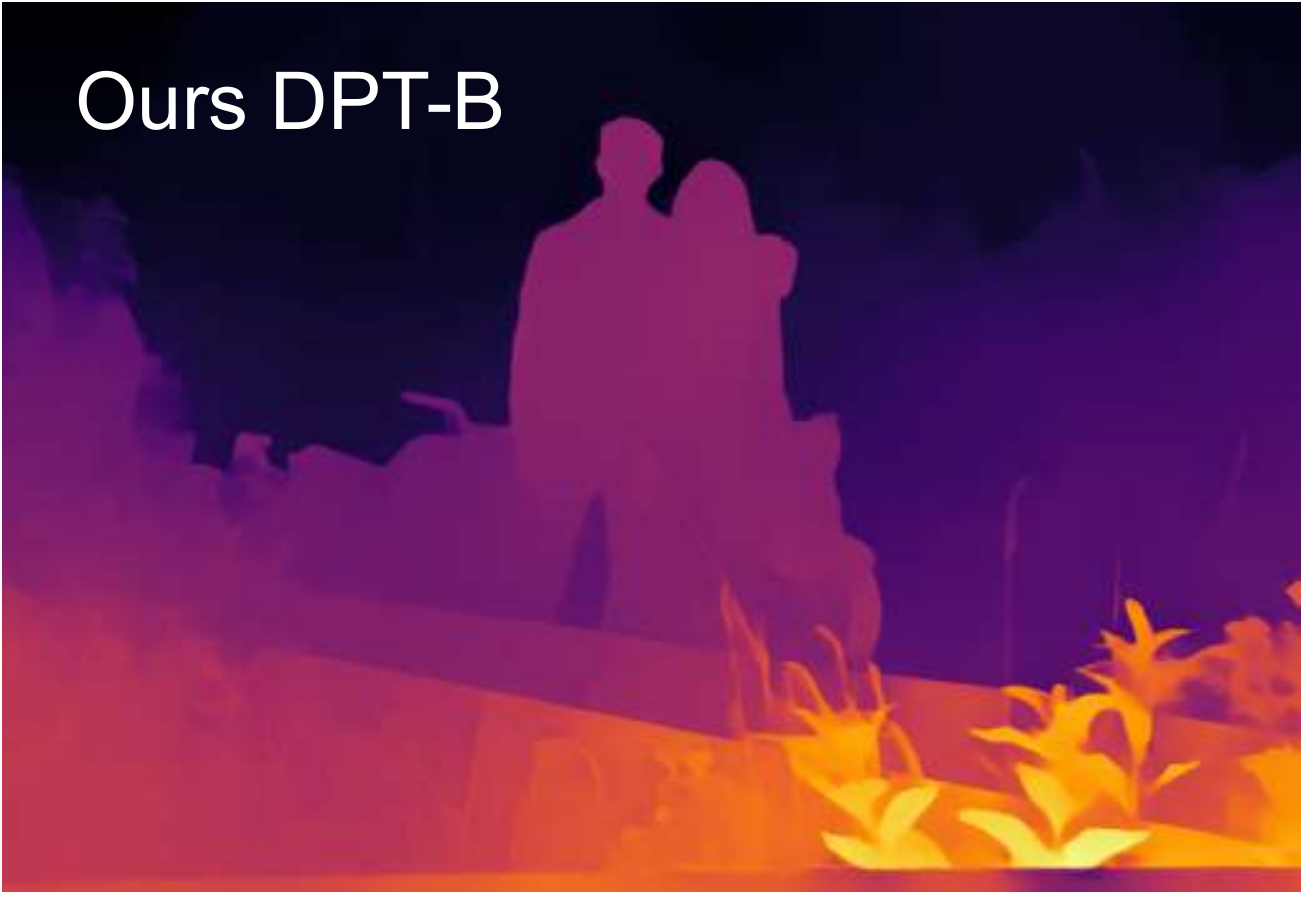}
  \includegraphics[width=0.48\linewidth, height=25mm]{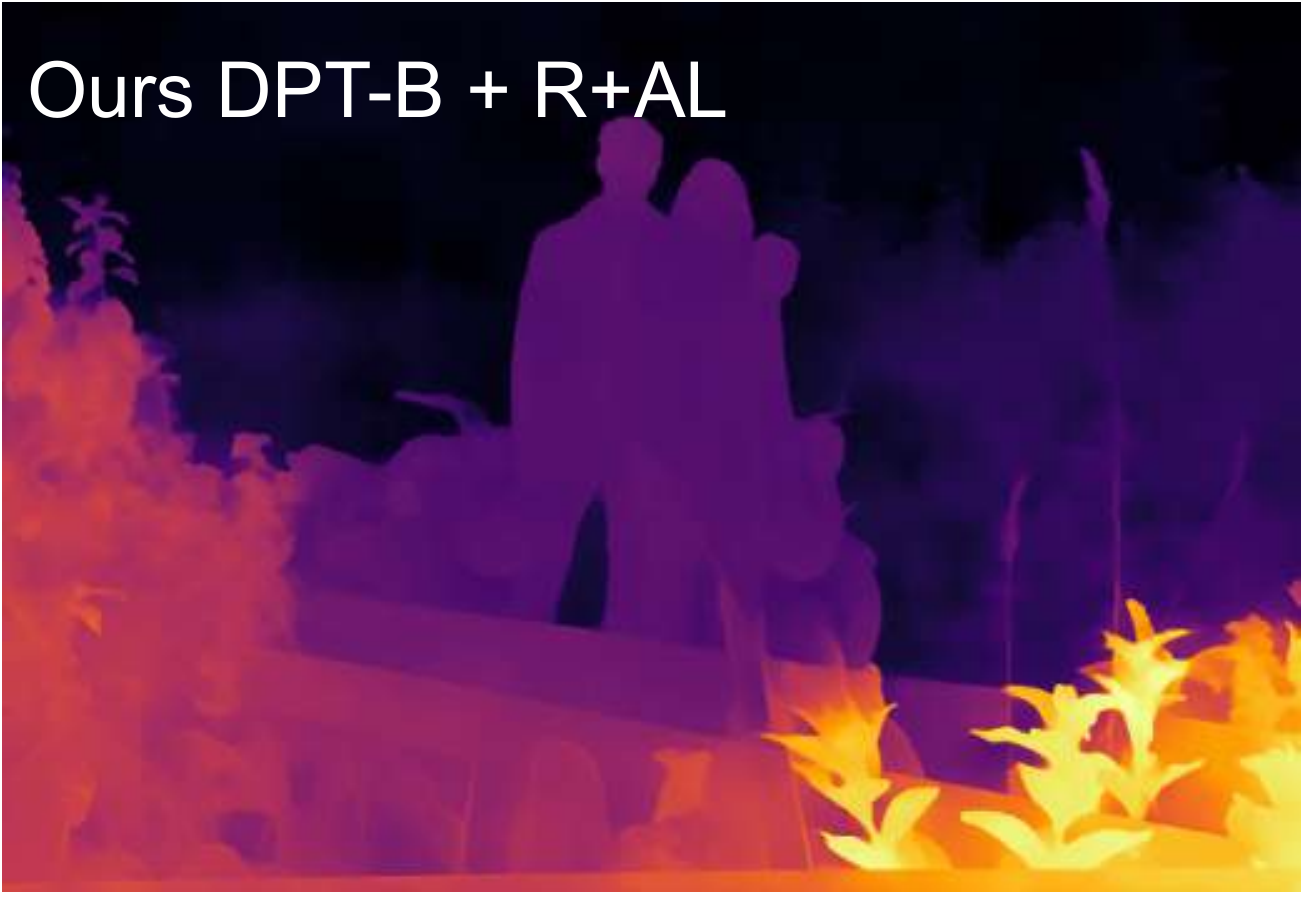}
  
  \end{subfigure}
\caption{Improvement over state-of-the-art DPT \cite{DPT}. Ours DPT-B and DPT-B+R+Al are two variants of DPT \cite {DPT} trained on the proposed HRSD datasts. The green rectangle represents area of the image where DPT fails to give precise and consistent depth compared to our algorithm.}
\label{introfig}
\end{figure}

Artificial intelligence's success in several computer vision applications has recently led to the low cost, small size, and wide applications of monocular cameras. Monocular depth estimation (MDE) algorithms are mainly based on neural networks and have shown great abilities in estimating depth from a single image \cite{monotransf,eigen2015predicting,monounsup,deepercnn,cnnmono,monofast}. These algorithms require leveraging high-level scene priors \cite{monolearn}, so training a deep neural network with supervised data becomes the defacto solution.

\begin{figure*}[h!]
\centering
\begin{subfigure}{.24\textwidth}
  \includegraphics[width=\linewidth, height=20mm]{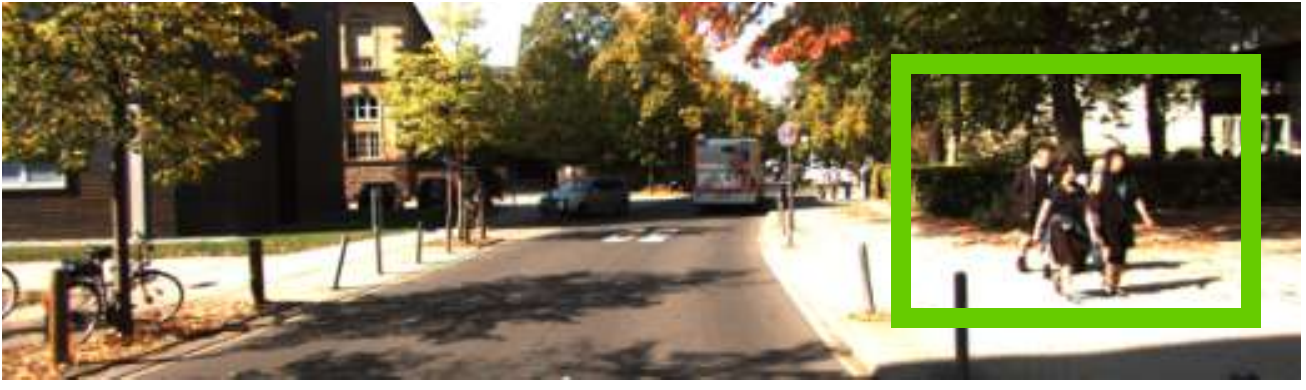}  \\
   \includegraphics[width=\linewidth, height=20mm]{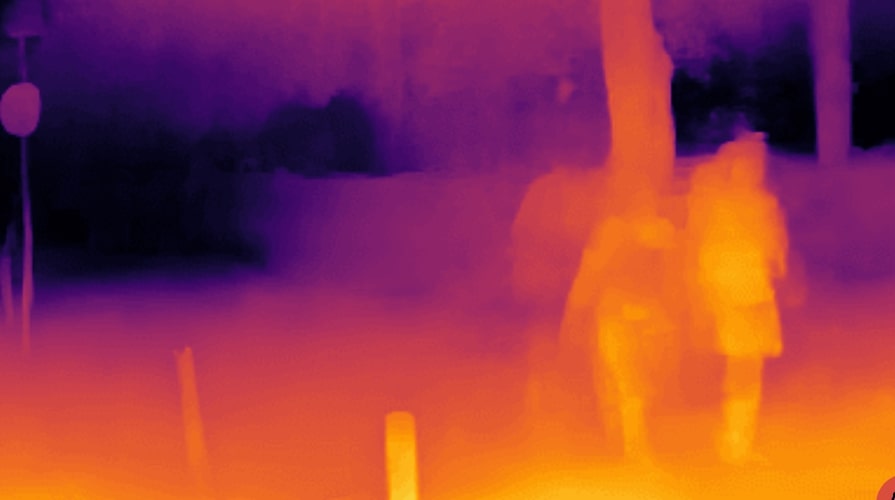} 
   \caption{KITTI RGB-D Dataset \cite{Kitti}}
\end{subfigure}
 \begin{subfigure}{.24\textwidth}
 \includegraphics[width=\linewidth,height=20mm]{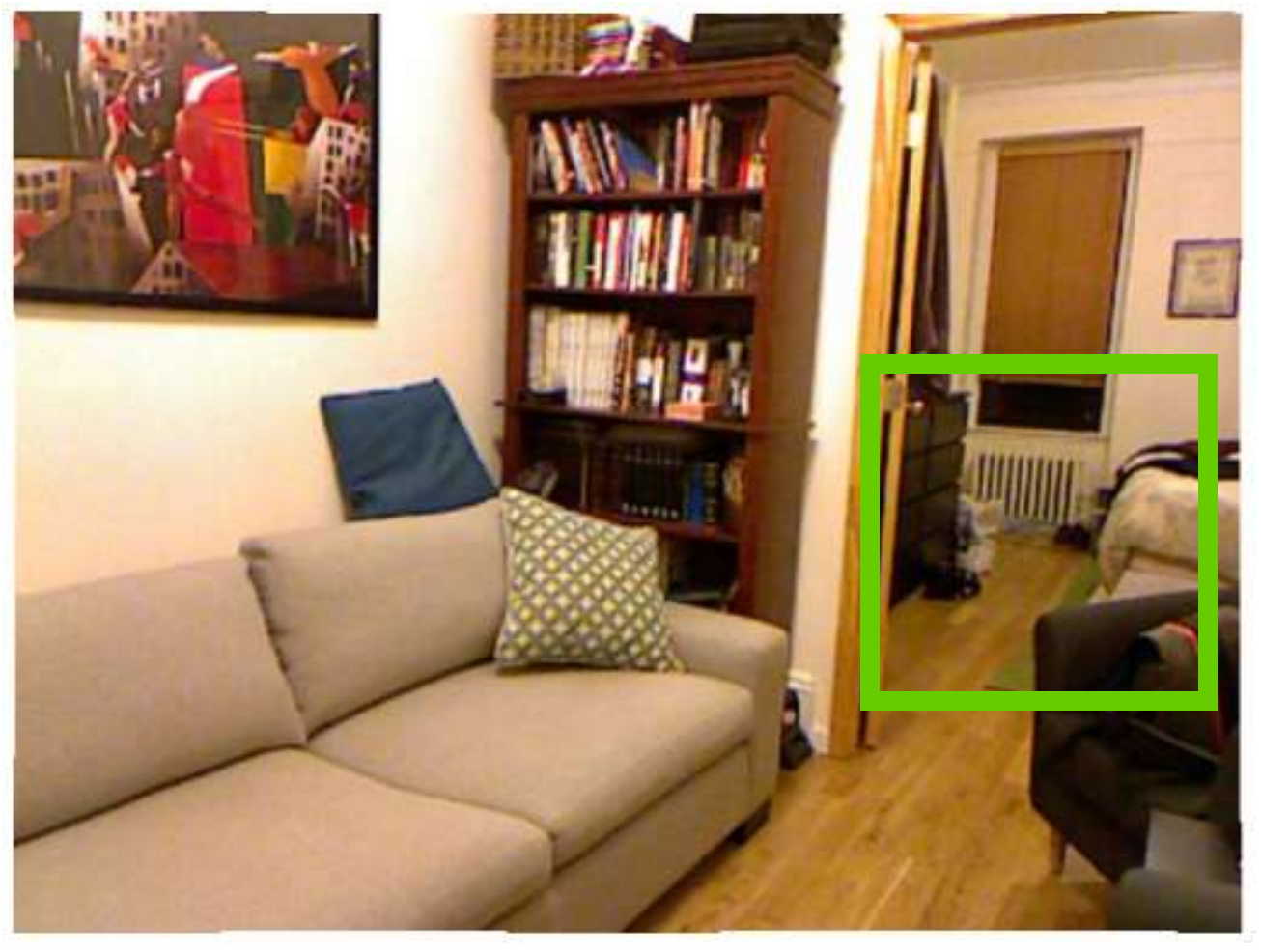}  \\
   \includegraphics[width=\linewidth, height=20mm]{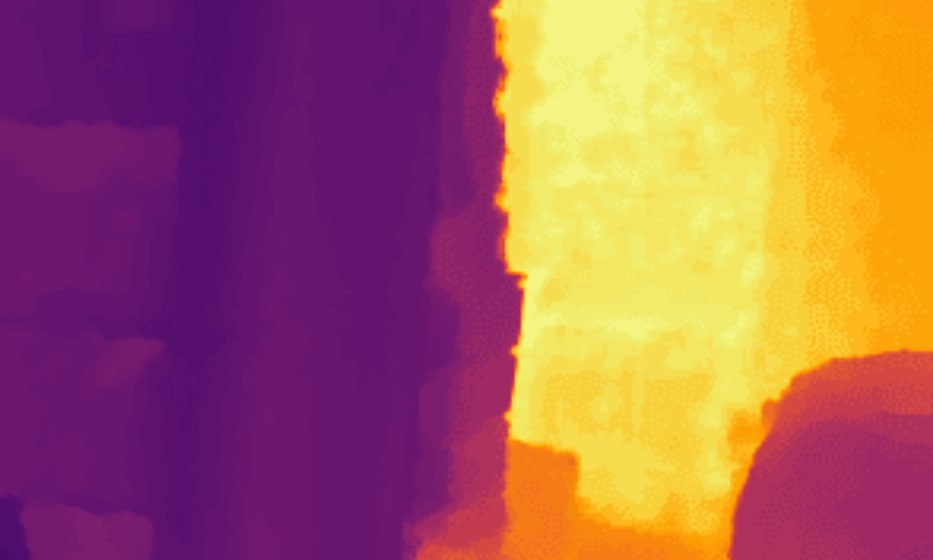} 
   \caption{NYU V2 Dataset \cite{NYU} }
\end{subfigure}
\begin{subfigure}{.24\textwidth}
  \includegraphics[width=\linewidth, height=20mm]{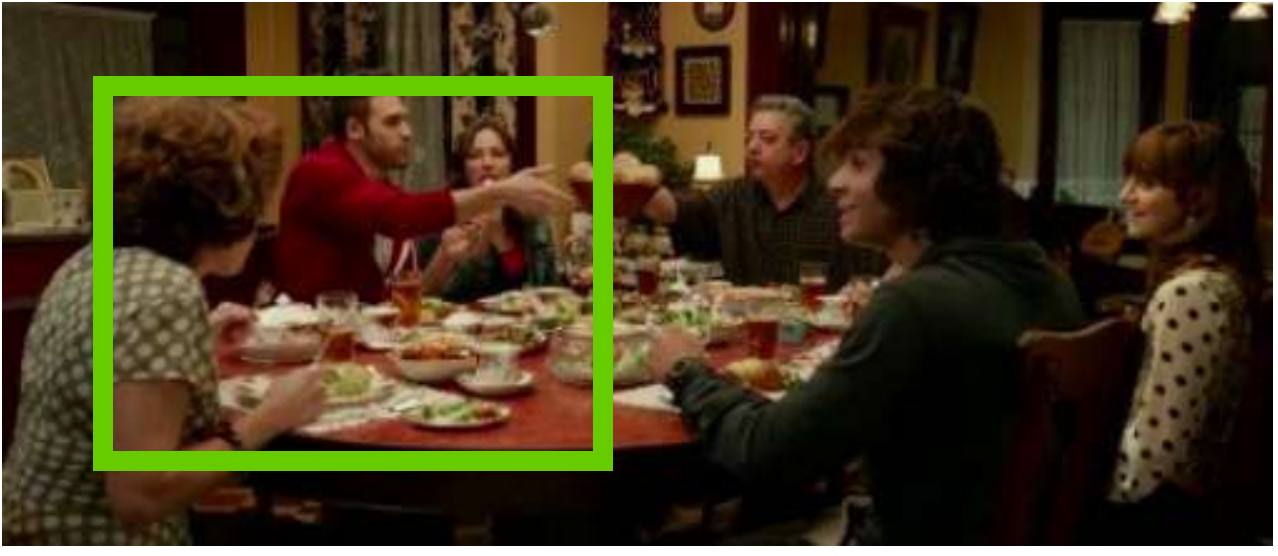}  \\
    \includegraphics[width=\linewidth , height=20mm]{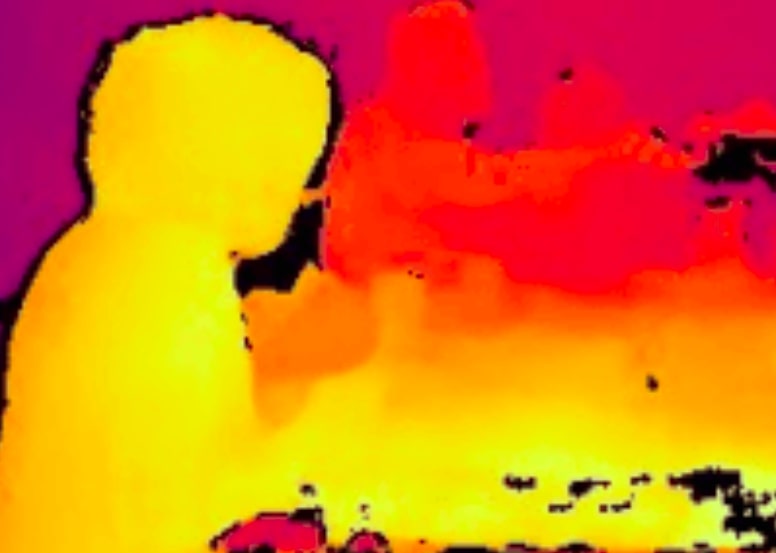} 
    \caption{MIDAS RGB-D \cite{midas}}
\end{subfigure}
\begin{subfigure}{.24\textwidth}
     \includegraphics[width=\linewidth, height=20mm]{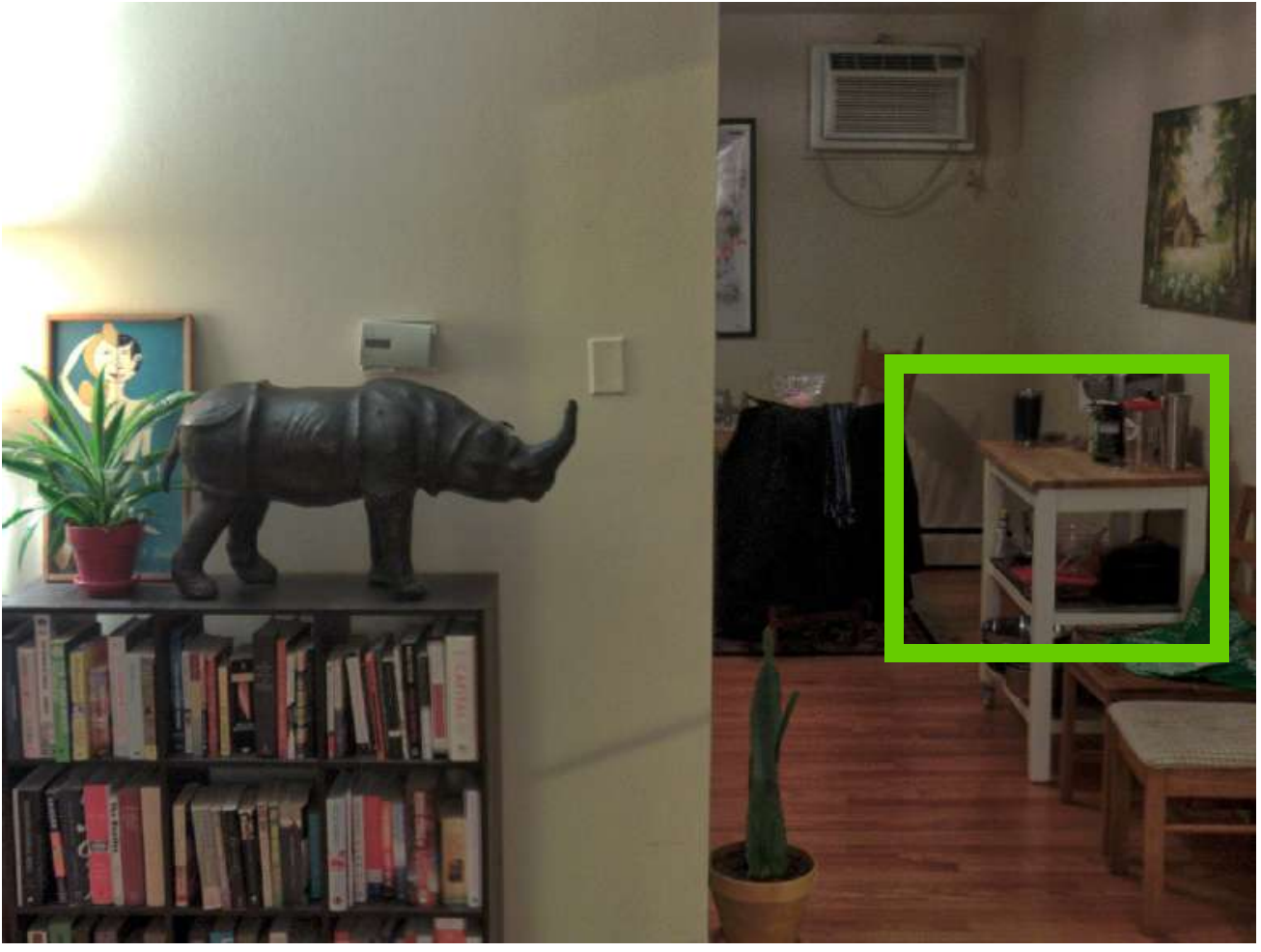} \\
    \includegraphics[width=\linewidth, height=20mm]{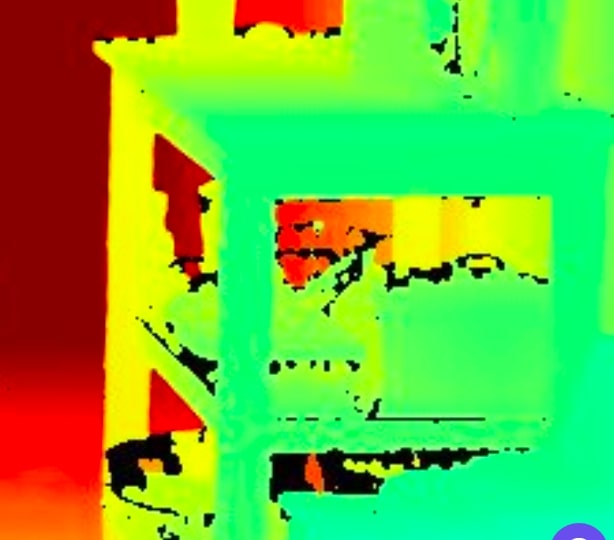}
    \caption{DioDE Dataset \cite{diode}}
\end{subfigure}
\caption{Example frames from publicly available datasets used for MDE. We zoom-in the problematic areas in the depth map. }
\label{existdata}
\end{figure*}
\par MDE algorithms using convolutional neural networks (CNN) deployed in an encoder-decoder structure learn a depth map with a similar spatial resolution to an RGB image. The encoder learns the feature representations from the input and provides a low-level output to the decoder. Using these features, the decoder first aggregates them and learns the final predictions. While CNN's have been the preferred architecture in computer vision, transformers have also recently gained traction motivated by their success in natural language processing \cite{attnall}. Transformer-based encoder structures  have significantly contributed to many vision-related problems, such as image segmentation \cite{setr}, optical flow estimation \cite{cotr}, and image restoration \cite{swin}. In contrast with CNN's that progressively down-sample the input image and lose feature resolution across the layers, vision transformer (ViT) \cite{ViT} processes feature maps at a constant resolution with a global receptive field at every stage. Feature resolution and granularity are important for dense depth estimation, and an ideal architecture should resolve features at or close to the resolution of the input RGB image.

Generally, an MDE algorithm based on CNN or transformer requires large RGB-D datasets consisting of diverse scenes with precise ground truth depth maps. However, the publicly available depth datasets mostly consist of ground-truth depth maps, either lower resolution (e.g., NYU V2) \cite{NYU} or sparse (e.g., KITTI RGB-D) \cite{Kitti}, or have been estimated from multiview depth estimation algorithms MiDaS\cite{midas}. Figure \ref{existdata} highlights some ground-truth depth maps from these publicly available datasets. This is one of the reasons why the existing MDE algorithms lack fine-grained details, as shown in Figure \ref{introfig}, and fail to perform accurately in diverse scenes. 
In this paper, we generate a high-resolution synthetic dataset using a commercial video game, Grand Theft Auto (GTA-V) \cite{gtav}. The dataset contains around 100,000 pairs of color images with precise dense depth maps of resolution $1920 \times 1080$, and we refer to this dataset as a High-Resolution Synthetic Depth (HRSD) dataset for monocular depth estimation. To show the effectiveness of the proposed HRSD datasets, we re-train the state-of-the-art transformer-based MDE algorithm: DPT \cite{DPT} with different experiments explained in section \ref{sec:train}. More specifically, we fine-tune DPT on the proposed HRSD datasets for dense depth prediction on high-resolution images, demonstrate the performance on other public datasets, and compare them with the SOTA algorithms. Further, to exploit the HRSD dataset, we propose adding a feature-extraction module with an attention-loss that further helps improve the results.\par    
In summary, we introduce the following concepts: 

\begin{itemize}
\item We have generated a high-resolution synthetic dataset (HRSD) from the game GTA-V \cite{gtav} with precise ground truth depth values. The HRSD dataset consists of diversified images enabling MDE networks to train on varied scenes, leading to a good generalization of real-world data. 
\item We propose to add a feature extraction module that processes the color image and converts them into feature maps to use them as patches for both the ViT \cite{ViT} and DPT algorithms \cite{DPT}. In addition, we optimize the training procedure by using an attention-based loss instead of  shift-invariant loss \cite{Playdepth, DPT}, improving performance in terms of efficiency and accuracy, resulting in smooth, consistent depth maps for high-resolution images.
\end{itemize}

We conduct experiments on standard public datasets with different input image sizes, such as NYU V2 \cite{NYU} ($640 \times 480$), KITTI \cite{Kitti} ($1216 \times 352$), and our HRSD ($1920 \times 1080$). We compare the performance with DPT \cite{DPT}, the state-of-the-art MDE algorithm, and Multi-res, \cite{multires} a depth estimation network built specifically for high-resolution images. We observe that the depth maps produced after training on the HRSD dataset outperformed existing algorithms on different public datasets.

\section{Related Work}
\textbf{RGB-D Dataset}
Various datasets have been proposed that are suitable for monocular depth estimation, i.e., they consist of RGB images with corresponding depth annotation of some form. These datasets differ in captured environments and objects, type of depth annotation (sparse/dense, absolute/relative depth), accuracy (laser, stereo, synthetic data), image resolution, camera settings, and dataset size. Earlier RGB-D datasets have relied on either Kinect \cite{kitti2,kitti1,NYU,sun-rgbd} or LIDAR \cite{scene,Kitti} or stereo vision \cite{middlebery} for depth annotation. Existing Kinect-based datasets are limited to indoor scenes; existing LIDAR-based datasets are biased towards scenes of man-made structures and have a low spatial resolution. Every dataset comes with its characteristics and has its own biases and problems \cite{rgbddata}.  

Ranftl et al. \cite{midas} introduced a vast  data source from 3D films capturing high-quality frames from movies to get a diversity of dynamic environments for depth estimation. They captured 80,000 frames at $1920 \times 1080$ resolution while using stereo matching to extract relative depth. However, stereo matching has its downside, as the depth maps are not precise and fail to perform on homogeneous scenes \cite{monodeep}. DIODE \cite{diode}, another high-resolution dataset generated using a laser sensor for dense depth annotation. It contained 25,000 indoor \& outdoor RGB images at a $768 \times 1024$, but it also led to inconsistent depth maps with artifacts in background objects and textured scenes, as shown in Figure \ref{existdata}. Synthetic dataset generation from computer games \cite{gtasemand,motsynth,hurl2019precise} has been used extensively for training computer vision algorithms like semantic segmentation \cite{gtaseman2,gtaseman1}, object detection \cite{gtaobject}, and depth estimation \cite{Playdepth, gtaflowdepth}. Mohammad et al.  \cite{Playdepth} have also used the GTA-V game to generate a synthetic RGB-D dataset but with relative depth at the focus. The resolution of the dataset used for training is $256 \times 256$ which is much smaller than the publicly available datasets. Furthermore, they need a preprocessing phase, such as histogram equalization, to use the datasets before feeding them to the training. For training, they use a Resnet architecture and process the RGB image and GT depth with resolution $256 \times 256$.

\par
\textbf{Depth from Single Image}
Convolutional networks within an encoder-decoder paradigm \cite{monotransf} is the standard prototype architecture for dense depth prediction from a single image. The building blocks of such a network consist of convolutional and sub-sampling as their core elements. However, CNN as an encoder suffers from a local receptive field problem \cite{banet}, leading to less global representation learning at higher resolutions. Several algorithms adapt different techniques to learn features at different resolutions to address this issue like dilated convolutions \cite{deeplab, hdeeplab} or parallel multi-scale feature aggression \cite{monoguide,multiscale}.

\par
Recently, transformer architectures such as vision transformer (ViT) \cite{ViT} or data-efficient image transformers (DeiT) \cite{Deit} have outperformed CNN architectures in image recognition \cite{ViT,Deit}, object detection \cite{DETR}, and semantic segmentation \cite{setr}. Inspired by the success of transformers in various topics, René et al. \cite{DPT} use a vision transformer for dense depth prediction and have outperformed all the existing MDE algorithms. Nonetheless, all transformer architectures are data-hungry networks \cite{transrobust,ViT} and thus require a huge dataset. 

\vspace{.5 cm}

\section{Proposed Method}
In this section, we introduce our high-resolution synthetic dataset (HRSD) and then discuss the proposed architecture changes to ViT \cite{ViT} and DPT \cite{DPT} algorithms to provide consistent and accurate dense depth maps on high-resolution images.
\par
\begin{figure*}[h!]
\begin{subfigure}{.98\textwidth}
  \includegraphics[width=0.16\linewidth,height=25mm]{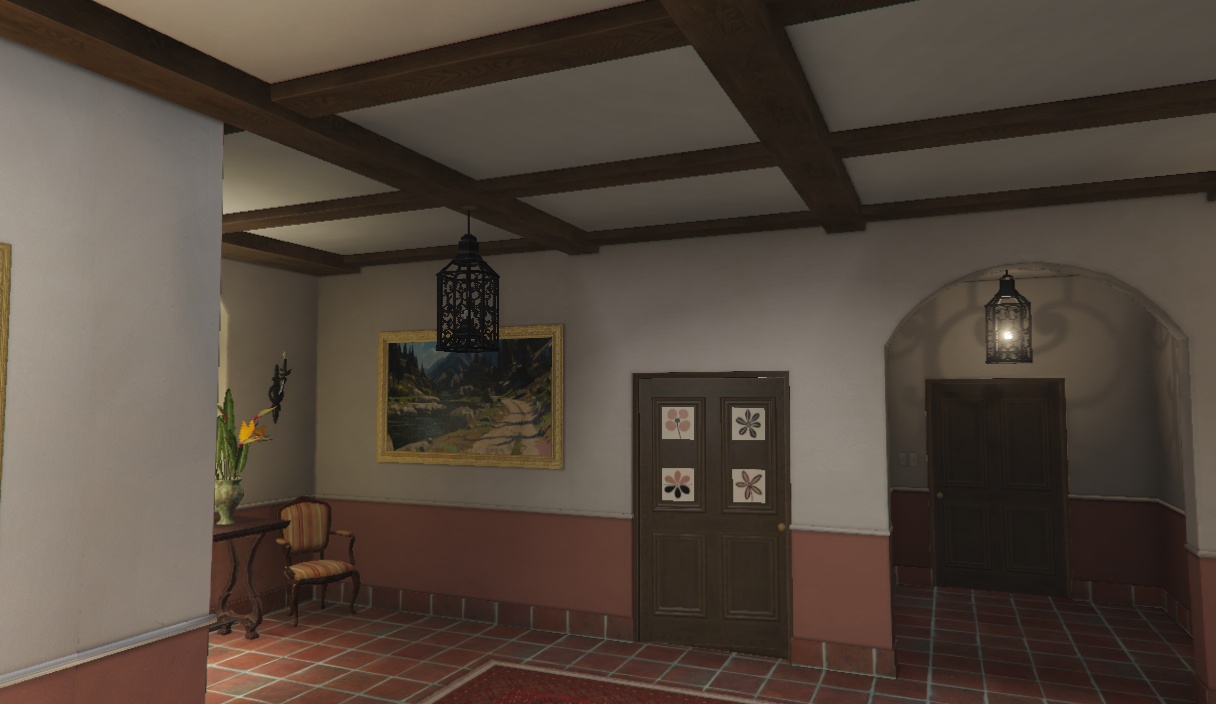}  
  \includegraphics[width=0.16\linewidth,height=25mm]{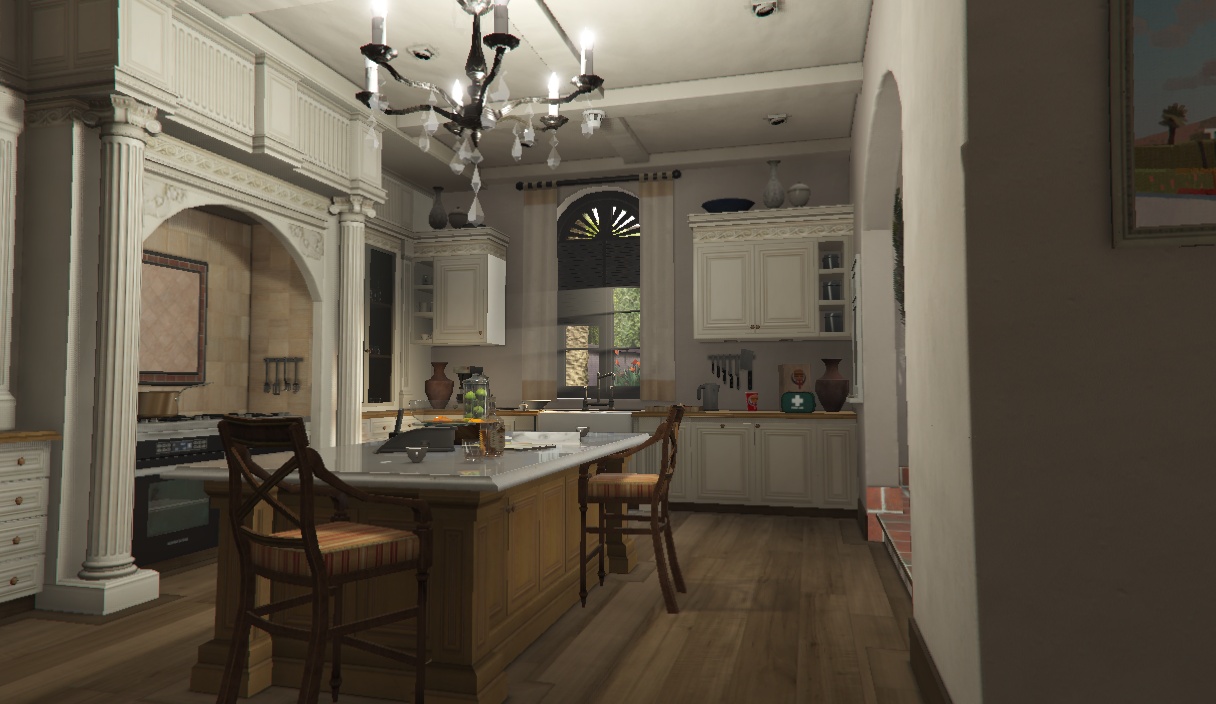}
  \includegraphics[width=0.16\linewidth,height=25mm]{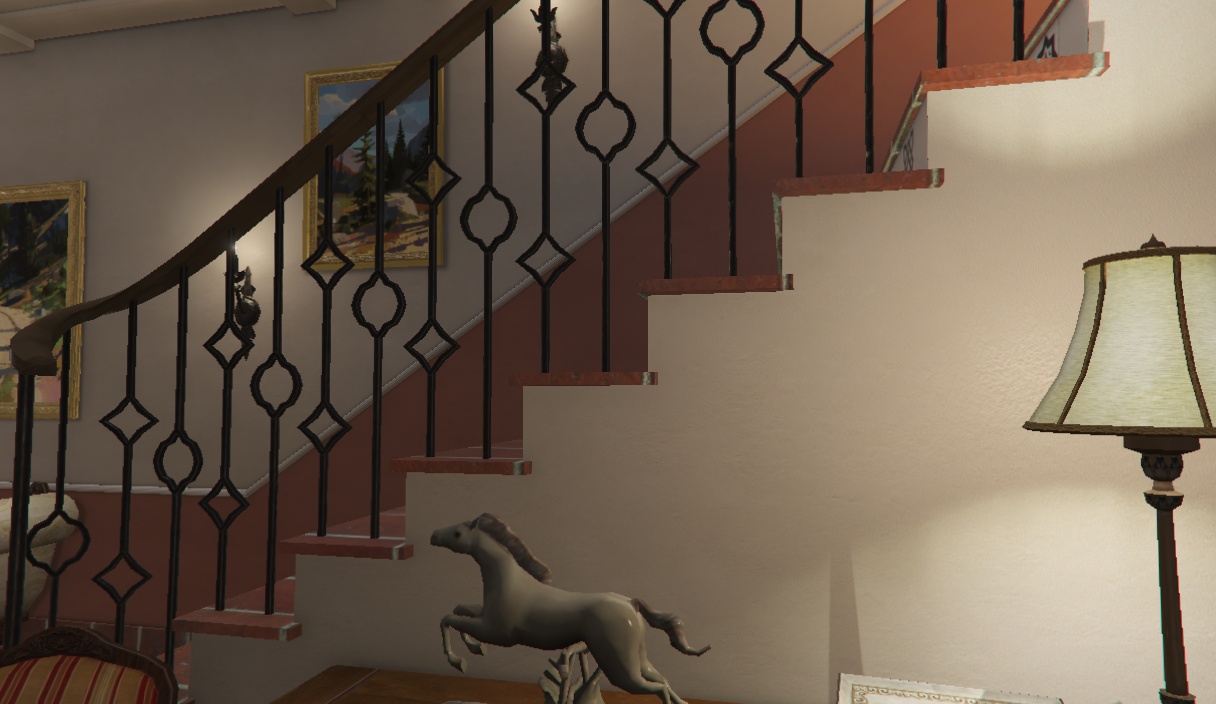}
     \includegraphics[width=0.16\linewidth,height=25mm]{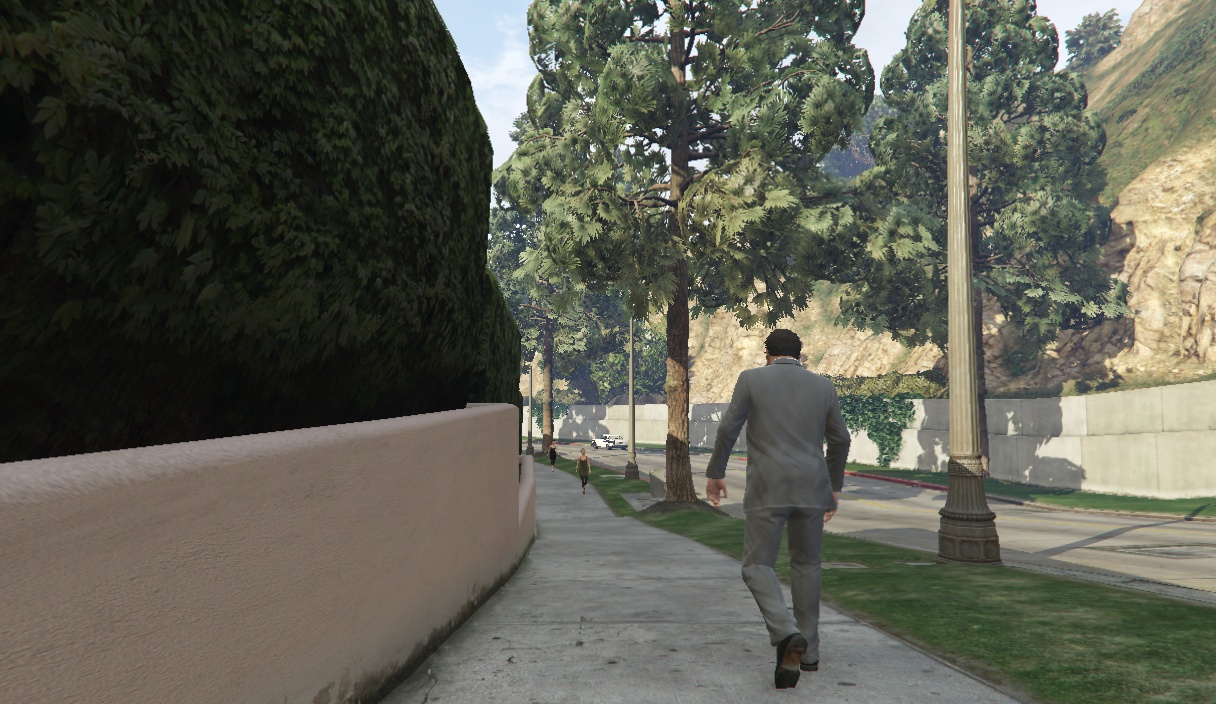}
      \includegraphics[width=0.16\linewidth,height=25mm]{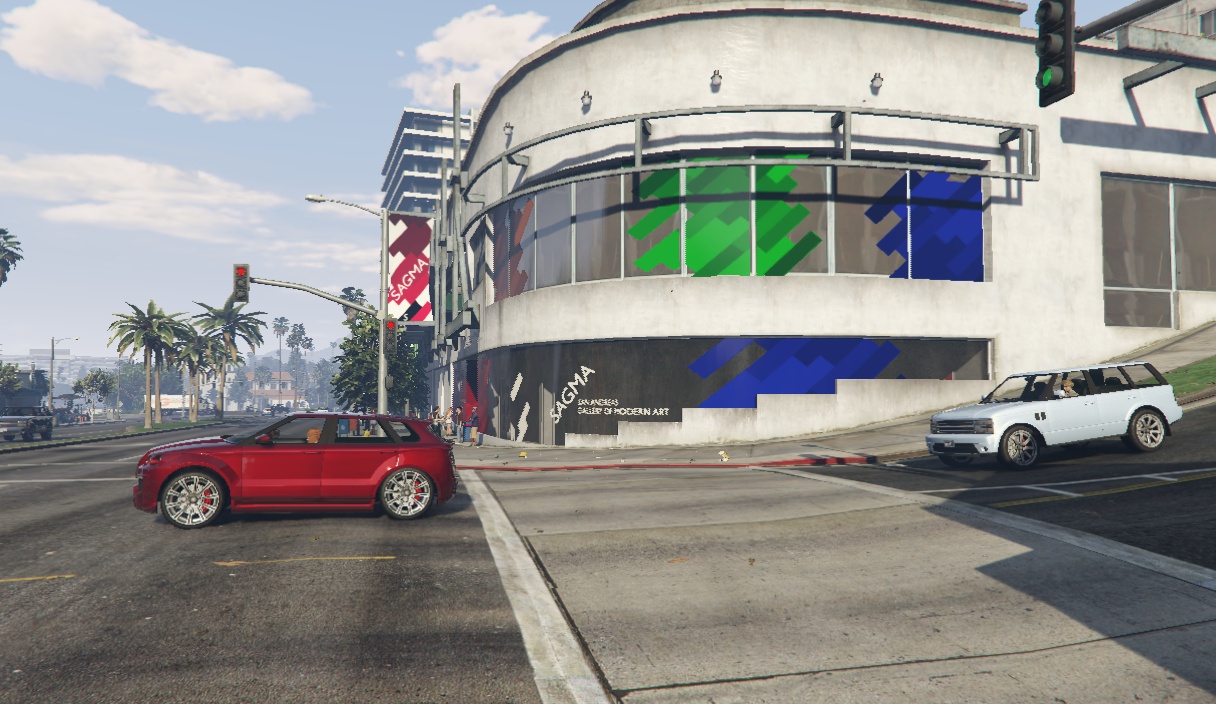}
        \includegraphics[width=0.16\linewidth,height=25mm]{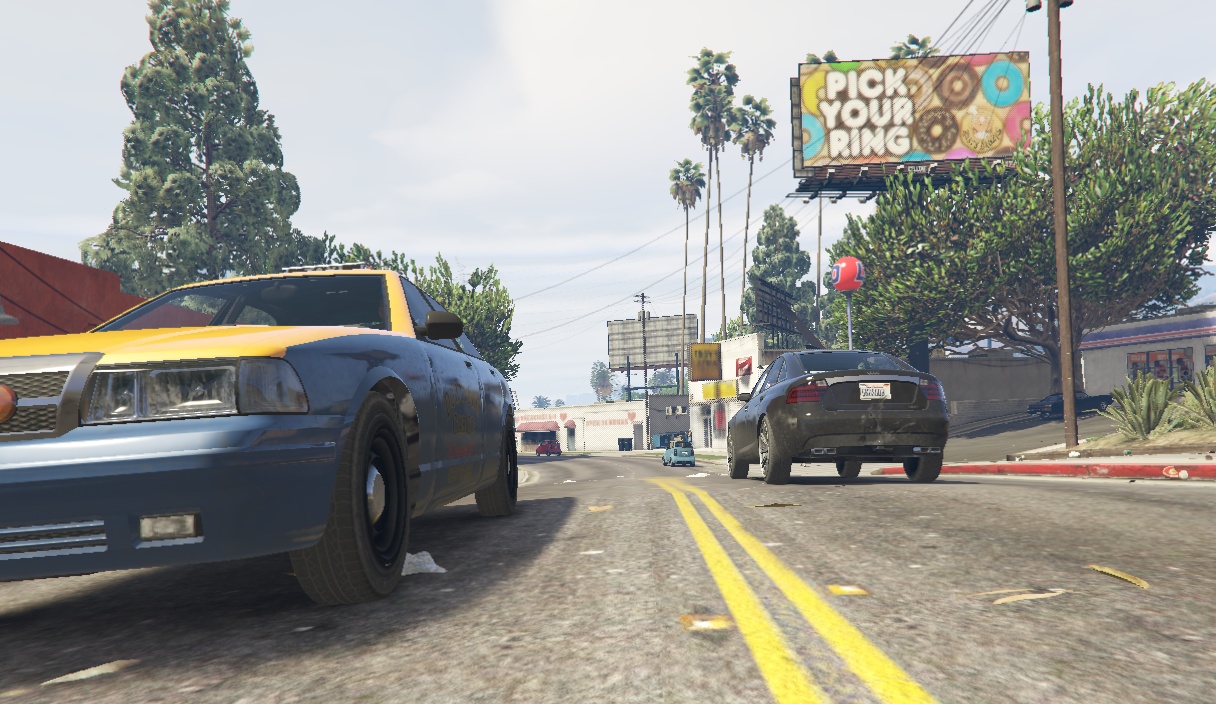}
  \end{subfigure}
\hfill
\begin{subfigure}{.98\textwidth}
  \includegraphics[width=0.16\linewidth,height=25mm]{images/GTA_depth/gta_indoor1.jpg}
\includegraphics[width=0.16\linewidth,height=25mm]{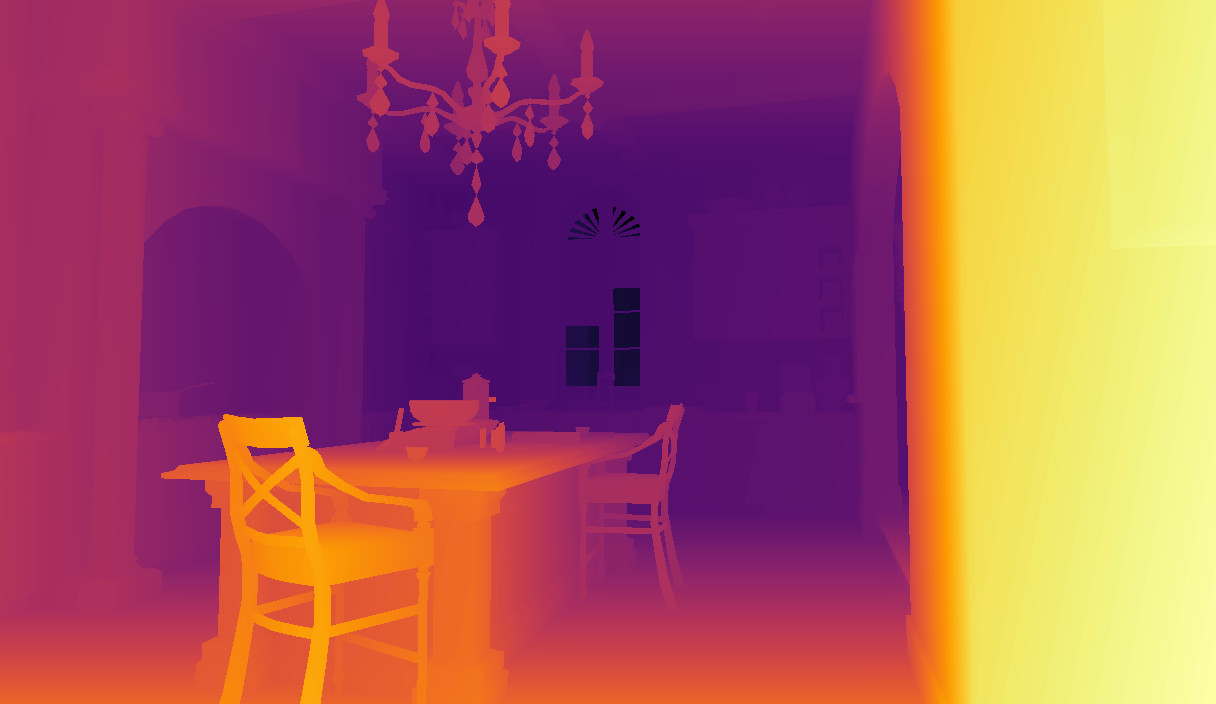}
\includegraphics[width=0.16\linewidth,height=25mm]{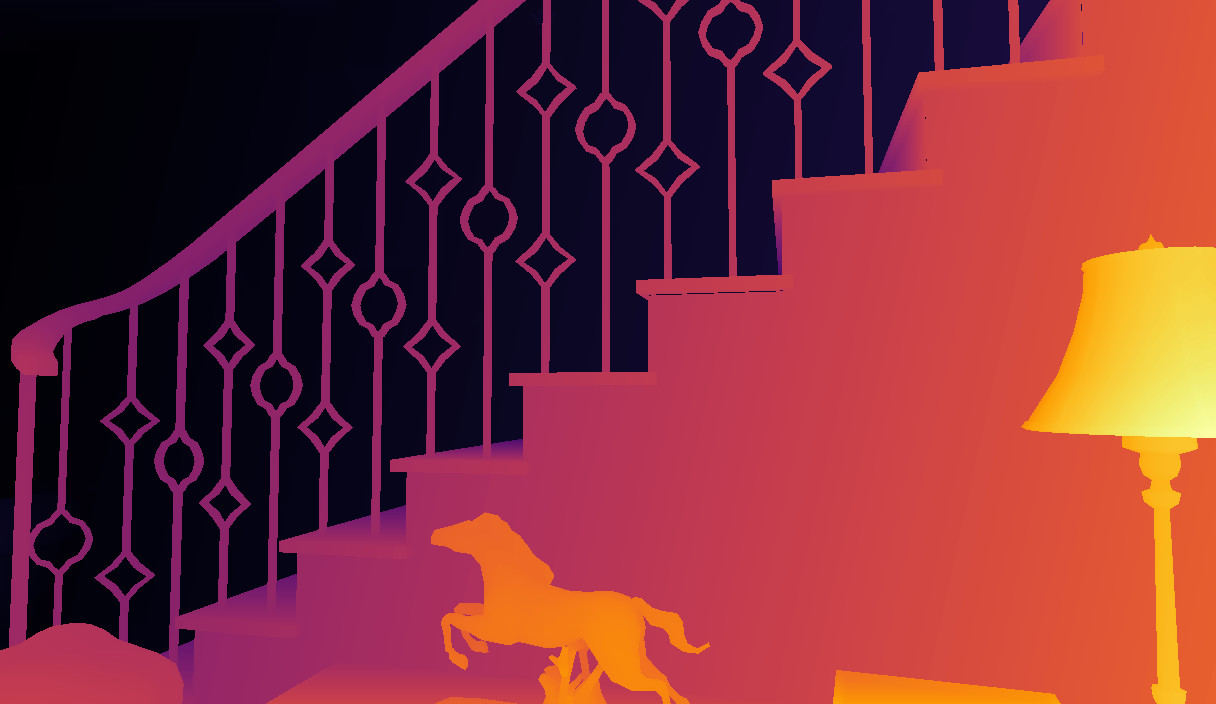}
\includegraphics[width=0.16\linewidth,height=25mm]{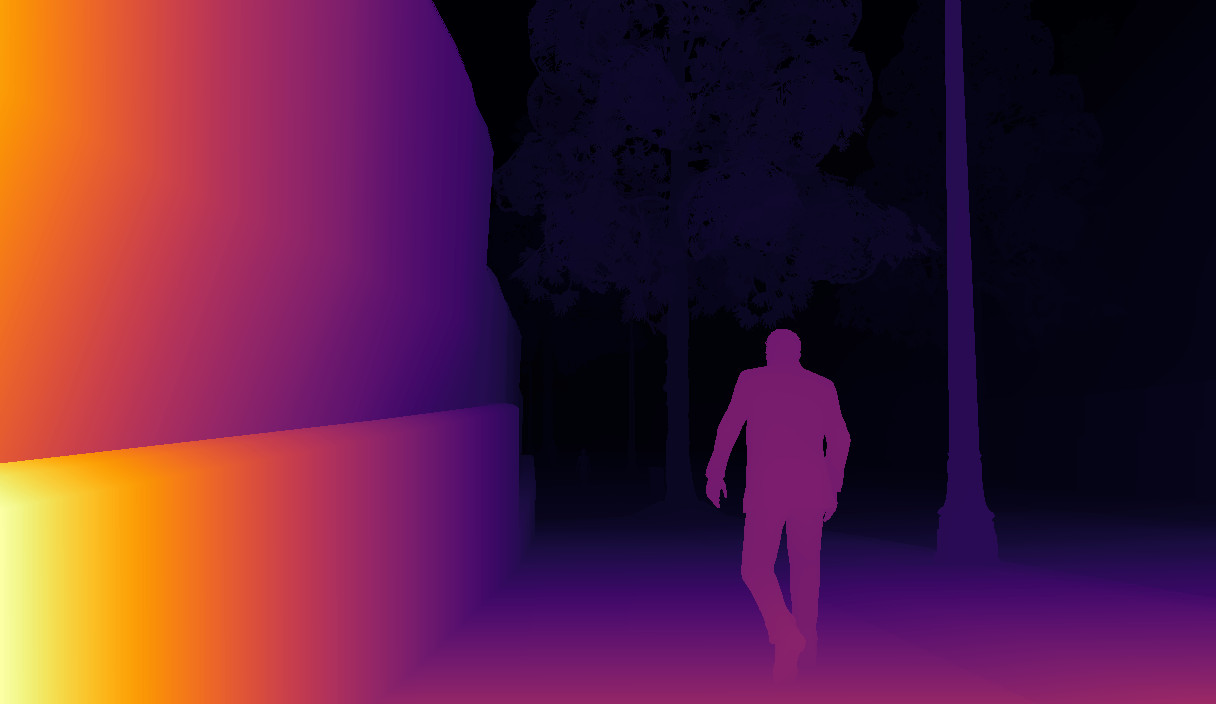}
  \includegraphics[width=0.16\linewidth,height=25mm]{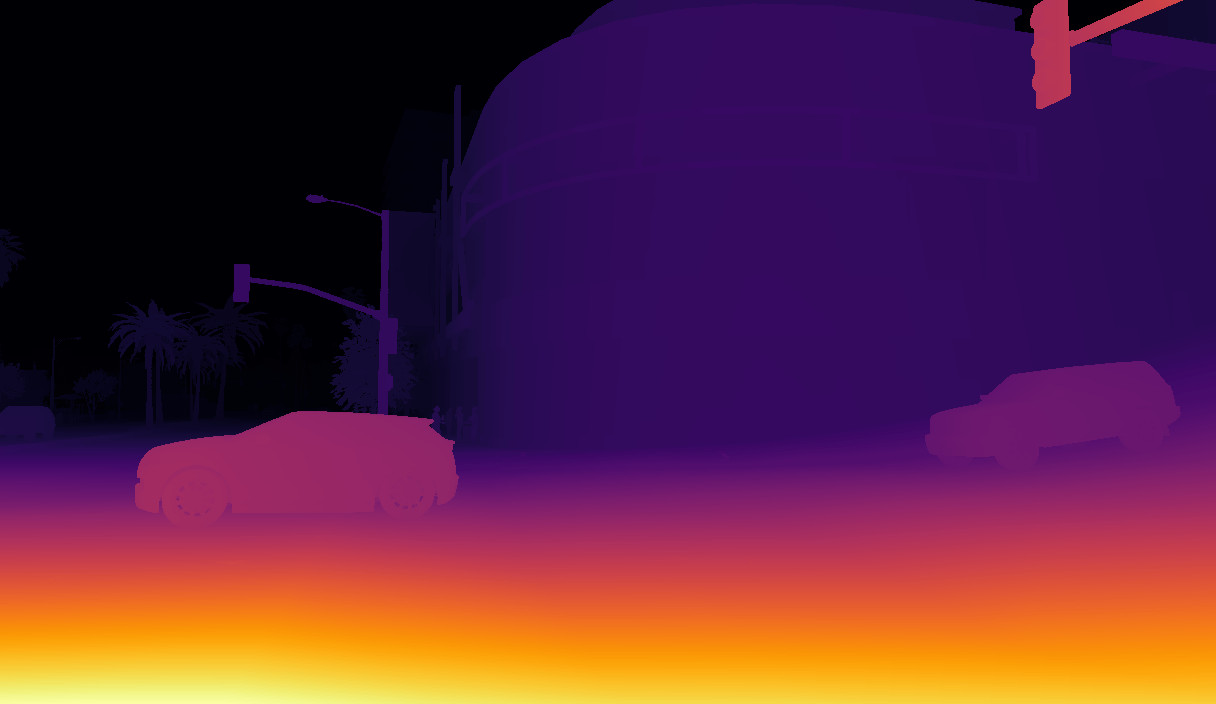}   
   \includegraphics[width=0.16\linewidth,height=25mm]{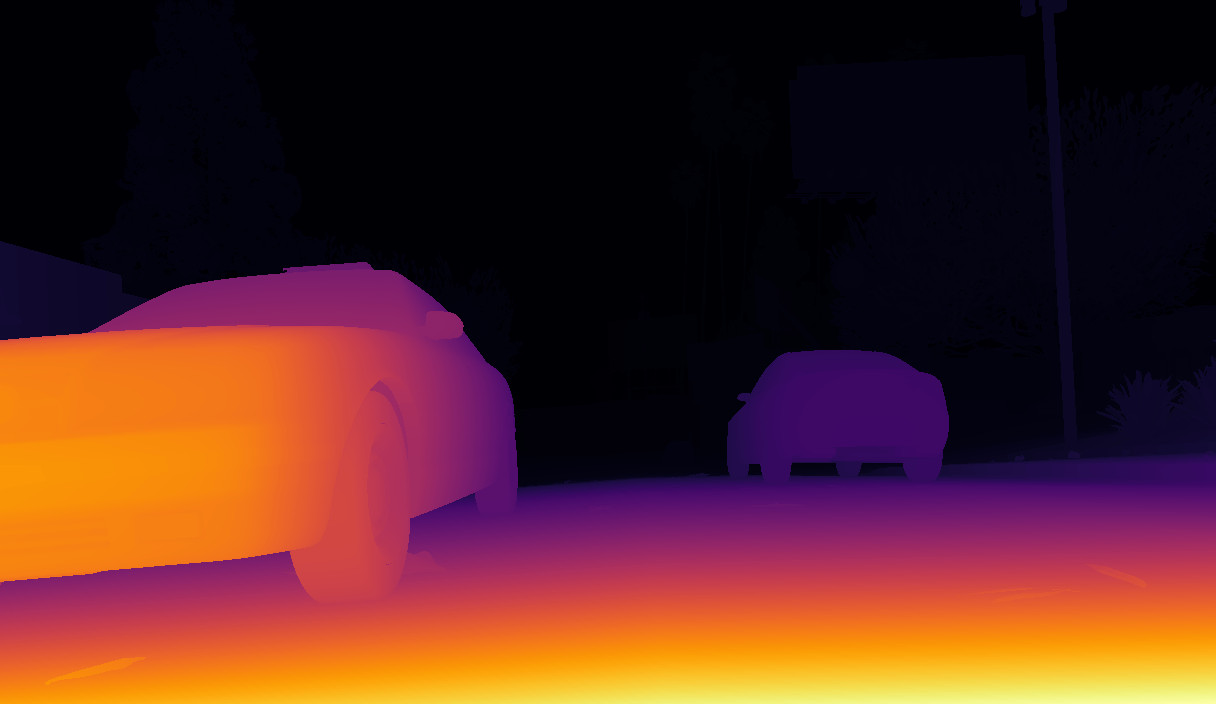}
\end{subfigure}
\caption{Examples from HRSD dataset consisting of indoor \& outdoor scenes with diversified objects and environments.}
\label{gtadepth}
\end{figure*}
\subsection{RGB-D Dataset}
Acquiring an accurate ground truth depth dataset for high-resolution images is challenging and expensive. Most RGB-D datasets have low image quality and sparse or relatively inaccurate depth maps.
 Inspired by the success of synthetic data in different computer vision applications \cite{gtaseman2, Playdepth,gtaobject, gtaseman1,gtaflowdepth}, we propose to generate a synthetic high-resolution RGB-D dataset for monocular depth estimation. The advantage of having a synthetic RGB-D dataset is to have precise ground truth depth maps for diverse color images. Also, one can control the lighting environments, objects, and background and the datasets can be as large depending on the applications.
\par

\par We use the Game Grand-Theft-Auto-V (GTA-V) \cite{gtav} to generate a high-resolution synthetic RGB-D dataset. On a high level, we use the GTA-V game’s in-build model and mechanics to calculate the ground truth (GT) depth and store them along with the high-resolution RGB image. Note that a similar idea of GTA-V-based synthetic data generation is designed for semantic segmentation \cite{gtasemand,gtaseman2} and depth estimation \cite{Playdepth,gtaseman1,gtaflowdepth}. Grand Theft Auto (GTA) \cite{gtav}, one of the prominent interactive games, consists of many diversified environments, including people, vehicles, recreational areas, and architecture. The precision  of graphics and 3D rendered models in this game is exceptional, making it a favorable alternative to acquiring demanding large-scale real-world datasets such as RGB-D. 

\par
\textbf{Real-time rendering}
To explain the data collection process, we must first review deferred shading, an important aspect of modern video games real-time rendering pipeline. Deferred shading utilizes geometric resources to produce depth and normal image buffers by communicating to the GPU \cite{real}. The intermediate outputs of the rendering pipeline are collected in buffers called G-buffers, which are then illuminated rather than the original scene. Rendering is accelerated significantly due to decoupled geometry processing, reflectance properties, and illumination. To utilize the rendering pipeline, we identify how the game communicates with the graphics hardware to extract the different types of resources, i.e., texture maps and 3D meshes. These resources are combined for the final scene composition. APIs such as OpenGL, Direct3D, or DirectX, provided via dynamically loaded libraries, are used for inter-communication. Video Games load these libraries into their application memory and use a wrapper to the specific library to initiate the communication and record all commands.  \par
\textbf{G2D and Scripthook V}
Using a DirectX driver, we communicate with GTA-V and redirect all rendering commands to the driver for extracting depth maps. To obtain the necessary image datasets under varying conditions from GTA-V, we use an image simulator software G2D: from GTA to Data \cite{g2d}. G2D allows to manipulate the environment on the fly, by injecting a customized mod in the game that controls global lighting and weather conditions by timing sunrises and sunsets. Using an accelerated time frame, we can create multiple high-resolution images from a single scene during sunrise, midday, sunset, and midnight. At the same time, automatic screen capture is enforced to record each frame displayed within the game, and the depth information is extracted. This enables us to accumulate a massive dataset as shown in Figure \ref{gtadepth}, comprising diverse environmental conditions (time of day, weather, season, etc.) and resulting in training computer vision algorithms that are robust and reliable in the real world.
\par
Using the above strategy, we can generate as many images, and in this paper, we synthetically generate $100,000$ images with resolution $1920 \times 1080$  along with depth maps. The minimum and maximum depths are set to $0.1$ m and $10$ m, respectively, for indoor scenes, and the max depth for outdoor scenes is $50$ m. We choose a small depth range to enable efficient training of our modified transformer network using $L_{1}$ loss instead of the scale-invariant loss \cite{eigen} used in most MDE networks \cite{adabins,Playdepth,DPT,midas}. We then use this dataset to train ViT \cite{ViT} \& the DPT \cite{DPT} algorithm for dense depth-map prediction. %for about 80, 000 different varied scenes at 1920x1080 resolution. We used.

\begin{figure}[t]
\centering
    \includegraphics[width=0.48\textwidth, height=0.29\textheight]{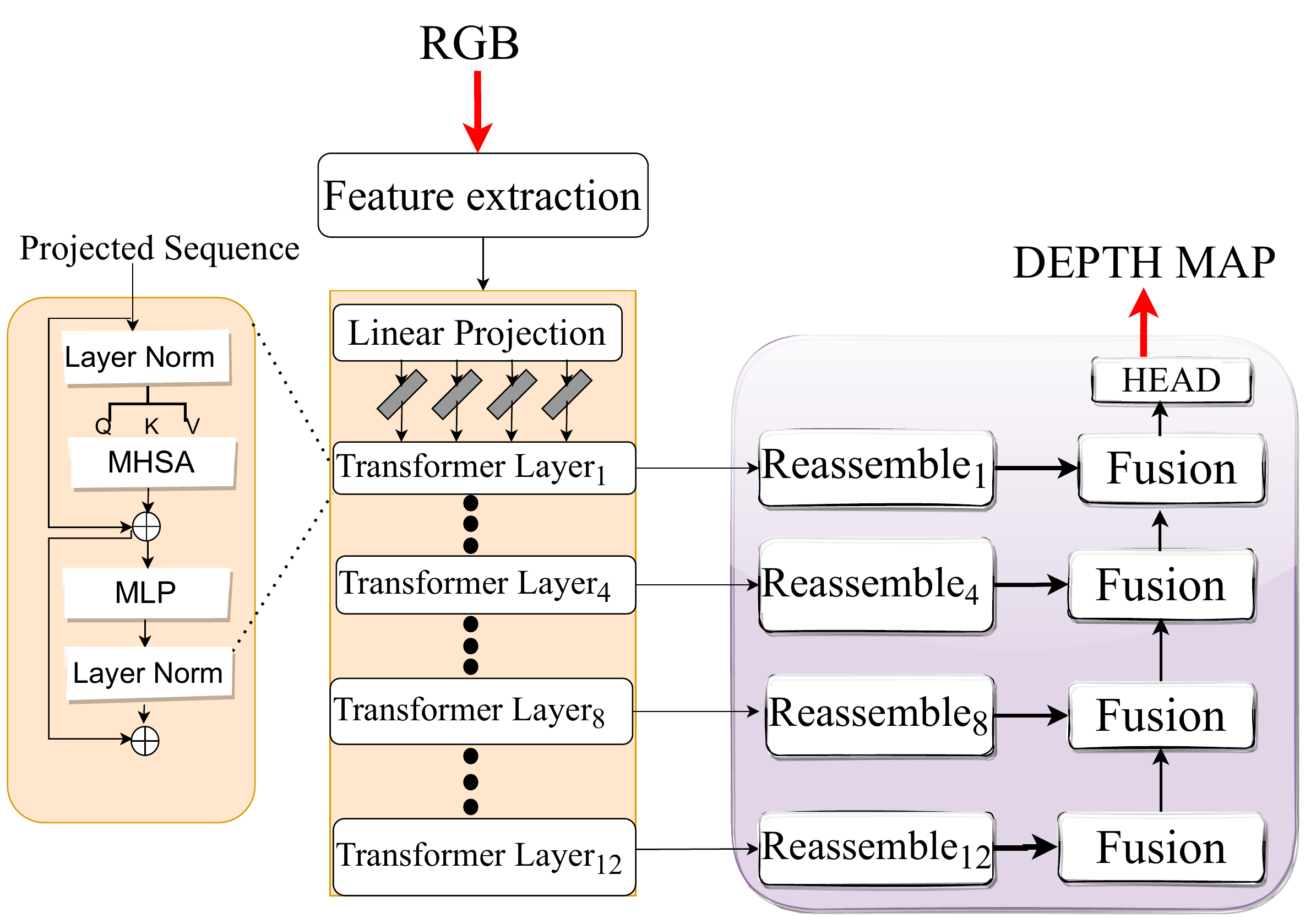}
    \caption{Overview of our proposed changes to DPT\cite{DPT} architecture. We introduce a feature extraction module.}
    \label{fig:overview}
\end{figure}

\subsection{Architecture}
\label{sec:arch}
 The state-of-the-art DPT \cite{DPT} algorithm uses ViT \cite{ViT} as a backbone encoder and then adds a decoder to get a depth map of the same resolution as a color image. Here, we discuss our modification to the DPT \cite{DPT} architecture. More specifically, we add a feature extractor module (pretrained Resnet \cite{resnet}) in the DPT \cite{DPT} architecture and a loss function consisting of attention supervision and $L_{1}$ loss to efficiently train the algorithm. The important blocks of the architecture are described below, and an overview of the network is depicted in Figure \ref{fig:overview}.
 
\textbf{Feature Extractor Module} Previous works \cite{medtrans,ViT,setr} split the input RGB Image 
$I \in \mathbb{R}^{H \times W \times 3} $
into equal size 2D patches and used linear projection to convert them into tokens for ViT. These tokens have a one-to-one correspondence with image patches and thus grow as the size of the input image increases. For high-resolution images, this creates a bottleneck for the ViT, given as the number of tokens increases, ViT's computation performance increases leading to higher inference time. As shown in Figure \ref{fig:overview}, we replaced the color-image patches in ViT \cite{ViT} (or DPT's encoder) with a feature extraction module similar to DPT-Hybrid \cite{DPT}. In this paper, we use pretrained ResNet \cite{resnet} for our feature extraction module and a detailed analysis is given in Table \ref{featurecomp}. 
\par 
The input sequence to the ViT now comes from a ResNet backbone \cite{resnet}. This feature module (ResNet-50 \cite{resnet}) converts the input image into patches of feature maps. We use the final layer feature map, which gives a pre-defined dimensional representation of any image size. To match the input dimension of the transformer with the output of the final ResNet-50 layer, we flatten the spatial dimension of these feature maps and project them to the transformer's dimension. So the vectorized patches from the output of Resnet are projected into a latent embedded sequence, the input to the first transformer layer, as shown in Figure \ref{fig:overview}. The input tokens are then processed by  $L$ transformer layers consisting of multi-headed self-attention (MHSA) and multi-layer perceptron (MLP) blocks. \par
\textbf{Vision Transformer Encoder} The spatial resolution of the initial embedding is maintained throughout the ViT encoder enabling a global receptive field at every transformer layer. This, along with MHSA being an inherently global operation, helps to achieve higher performance on high-resolution images. 
We have experimented with different values of $L$, with $12$ transformer layers providing consistent dense depth maps similar to DPT-Base \cite{DPT}. At each layer $L$, the input of MHSA is a triplet of $Q$ (query), $K$ (key), and $V$ (Value) computed as follows. \cite{ViT, attnall}
\begin{equation}
Q=z_{\ell-1} \times W_{Q}, K=z_{\ell-1} \times W_{K}, V=z_{\ell-1} \times W_{V}
\end{equation}
Here, $z_{l-1}$ is the output from the previous transformer layer, with $z_{0}$ being the input to the first layer. Then, the attention head ($AH$) is calculated using the triplet $Q$, $K$, and $V$ as given in \cite{ViT, attnall}. Later, all the $AH$'s are combined with a weight matrix calculated during training for multi-head attention. The MHSA output is then fed to the MLP block, which calculates the final output.

\par 
\textbf{Convolutional Decoder}. Our decoder resembles the one used in DPT \cite{DPT} as it assembles the set of tokens from different transformer layers at various resolutions. An image-like representation is recovered from the output of the encoder using a three-function calculation called the reassemble operation \cite{DPT}. We use four reassemble blocks which extract tokens from the $1^{\text {st}}$, $4^{\text {th }}$, $8^{\text {th }}$, $12^{\text {th }}$ (final) transformer layers. Transformer networks need more channels than their convolutional counterparts, so we double the number of channels in the three last reassemble modules. Each stage in the decoder layer, known as fusion blocks, is based on refinement \cite{refinenet}. They progressively combine the feature maps from consecutive stages into the final dense prediction. Unlike in the DPT decoder, batch normalization is helpful for dense depth prediction. We also reduce the number of channels in the fusion block to 96 from 256 in DPT \cite{DPT} to enable faster computations. The final block is the head block which outputs relative depth for each pixel.
\subsection{Attention-Based Loss}

The depth range of our HRSD dataset allows us to utilize the standard loss function for depth regression problems known as $L_{1}$ loss or Mean Absolute Error loss (MAE). Unlike a scale-invariant loss \cite{eigen}, whose variants are used for many MDE networks \cite{adabins,Playdepth,DPT,midas}, $L_{1}$ loss is more efficient and performs better \cite{l1loss}. However, training with only $L_{1}$ loss leads to discontinuities and noisy artifacts in depth maps \cite{encdecmono}. Other loss functions considered  to aid $L_{1}$ loss include employing the edge accuracy between real and predicted depth maps known as Structural Similarity (SSIM) \cite{ssim} used for the MDE network \cite{monotransf}. Inspired by \cite{chang2021transformer}, in our method, we use an attention-based supervision loss to smooth the overall prediction and control the number of depth discontinuities and noisy artifacts in the final output. 
\begin{table}[t!]
\centering
\renewcommand{\arraystretch}{1.0}
\setlength{\tabcolsep}{2.5pt}

\begin{tabular}{|c|c|c|c|c|}\hline
\multirow{3}{*}{\textbf{Dataset}} & \multirow{3}{*}{\textbf{Algorithms}} & \multicolumn{3}{c|}{\textbf{ Error \& Accuracy}} \\
\cline{3-5}
&  & \textbf{AbsRel}  $\downarrow$ & \textbf{RMSE} $\downarrow$  & $\boldsymbol{\delta < 1.25}$ $\uparrow$ \\ 
\hline 
\multirow{6}{*}{\rotatebox[origin=c]{90}{ \textbf{NYU V2} \cite{NYU}}} & ViT-B  & 0.115 & 0.509 & 0.828 \\
 
& ViT-B + R  & 0.108 & 0.416 & 0.875 \\
 
& ViT-B + R + AL  & \textbf{0.104} & \textbf{0.362} & \textbf{0.916}\\
\cline{2-5}
& DPT-B & 0.101 & 0.375 & 0.895  \\
 
& DPT-B + R  & 0.103 & 0.364  &  0.903 \\
 
& DPT-B + R + AL  & \textbf{0.094} & \textbf{0.310} & \textbf{0.945}  \\

\hline
\multirow{6}{*}{\rotatebox[origin=c]{90}{ \textbf{KITTI} \cite{Kitti}}} & ViT-B   & 0.106 & 4.699 & 0.861 \\
 
& ViT-B + R  & 0.101 & 4.321  & 0.889 \\
 
& ViT-B + R + AL  & \textbf{0.078} & \textbf{2.933} & \textbf{0.915} \\
\cline{2-5}
& DPT-B & 0.098 & 3.821 & 0.894 \\
 
& DPT-B + R  & 0.069 & 2.781 & 0.939 \\
 
& DPT-B + R + AL & \textbf{0.056} & \textbf{2.453}  & \textbf{0.962} \\
 
\hline
\multirow{6}{*}{\rotatebox[origin=c]{90}{ \textbf{HRSD}}} & ViT-B  & 0.125 & 0.471 & 0.828 \\
 
& ViT-B + R  & 0.107 & 0.342 & 0.882  \\
 
& ViT-B+ R + AL & \textbf{0.099} & \textbf{0.322} & \textbf{0.912} \\
\cline{2-5}
& DPT-B & 0.118 & 0.421 & 0.835 \\
 
& DPT-B + R  & 0.101 &  0.330  & 0.894 \\
 
& DPT-B + R + AL  & \textbf{0.074} & \textbf{0.288} & \textbf{0.921} \\
\hline
\end{tabular}
\caption{Quantitative comparison on three RGB-D datasets. The three variants of ViT \cite{ViT} and three variants of DPT \cite{DPT} are trained on the proposed HRSD datasets.}
\label{intercomp}
\end{table}

To estimate the true values of the attention map at each pixel $p$,  we calculate ${A}_{p}$ from the ground-truth depth map as 
\begin{equation}
{A}_{p}=\operatorname{SOFTMAX}\left(-\lambda\left|y_{p}-\hat{y}_{p}\right|\right)
\end{equation}
where $y_{p}$, $\hat{y}_{p}$ are the ground-truth and predicted depth map, respectively, and $\lambda$ is the hyper-parameter. We use the ground truth attention map ($A_{p}$) and the predicted attention values $(\hat{A}_{p})$ to calculate the attention-based loss term \cite{chang2021transformer}. 
\begin{equation}
\mathcal{L}_{\text {as}}= \frac{1}{n}\sum_{p=0}^{n}\left|A_{p}-\hat{A}_{p}\right|
\end{equation}

\par
For training our final network, we define the final loss $L$ between $y$ and $\hat{y}$ as the weighted sum of two loss terms:-

\begin{equation}
L(y, \hat{y})=\lambda L_{\operatorname{depth}(}(y, \hat{y})+L_{as}(y, \hat{y})
\end{equation}
where $L_{depth}$ is the point-wise $L1$ loss defined on depth values and is given as.
\begin{equation}
L_{\operatorname{depth}}(y, \hat{y})=\frac{1}{n} \sum_{p=0}^{n}\left|y_{p}-\hat{y}_{p}\right|
\end{equation}
Note that only one weight parameter $\lambda$ is required for calculating the loss function, and empirically $\lambda$ = 0.1 is set in equation (4)  \cite{chang2021transformer}.

\section{Experiments}

In this section, we make an extensive study that demonstrates the advantage of the proposed HRSD dataset. We evaluate public datasets, such as KITTI \cite{Kitti} and NYU \cite{NYU}, and the HRSD dataset. We show quantitative and qualitative comparisons in our experiments for analysis and discussion.

\subsection{Training}
\label{sec:train}
\textbf{Architecture details} DPT \cite{DPT} uses an encoder-decoder architecture. It uses ViT \cite{ViT} architecture for its encoder and designs a decoder to get a depth map of the same size as the color image. Thus, DPT \cite{DPT} use pretrained ViT \cite{ViT} weights as their initial encoder weights to train the entire encoder-decoder architecture on datasets, such as KITTI (outdoor) \cite{Kitti}, NYU V2 (indoor) scenes \cite{NYU}, and the MIDAS 3D dataset \cite{midas} for dense depth prediction. 
% For training, we use DPT \cite{DPT} architecture
We adopt the same transformer-based encoder-decoder architecture used in DPT for our experiments and improve it further, as mentioned above in section \ref{sec:arch}. We initialize the encoder weights with DPT's \cite{DPT} and train the entire encoder-decoder architecture on the proposed HRSD datasets (the decoder is initialized from scratch). To show the effect of proposed architecture changes, such as feature module and attention loss, we do six training experiments on the proposed HRSD dataset. Three training is done using DPT encoder weights \cite{DPT}, and they are:

\emph{1)} Training with DPT \cite{DPT} encoder weights \footnote{In this paper, for training and evaluation, DPT weights refers to DPT-Hybrid weights as provided in the original DPT \cite{DPT} paper.} on the proposed HRSD dataset with no changes in architecture and loss functions which refer to DPT-B in this paper. \emph{2)} Training DPT \cite{DPT} weights with feature module (DPT-B + R). \emph{3)} Training DPT \cite{DPT} with feature module and attention-loss (DPT-B + R + AL). 

Later, we do a similar process using ViT weights \cite{ViT}, i.e., initialize the encoder weights with ViT's \cite{ViT} and train the entire encoder-decoder architecture on the proposed HRSD datasets and do three separate trainings.

\textbf{Training details}
 Our proposed algorithm is implemented in PyTorch, and we use $4$ NVIDIA RTX A6000 48GB GPU for training.  
  The proposed HRSD dataset contains $100,000$ images of resolution $1920 \times 1080$ and we use $75,000$ for training, $15000$ for validation, and $10000$ for the test dataset. We crop the images to the nearest 32 multiples, and the network outputs the depth at the same resolution as of color image. We trained the model for $80$ epochs with a batch size of $4$ and use the ADAMW \cite{adamw} optimizer with $\beta$1 = 0.9 \& $\beta$2 = 0.999 and a learning rate of \num{1e-4} for encoder \& \num{1e-5} for the decoder. The learning rate decayed after $15$ epochs by a factor of $10$. 
Finally, we test the algorithm's performance on different datasets, such as KITTI, NYU, and HRSD, for indoor and outdoor scenes to show the generalization and robustness of the proposed algorithm in real-world scenes. 

\begin{table}[t!]
\centering
\renewcommand{\arraystretch}{1.0}
\setlength{\tabcolsep}{2.5pt}

\begin{tabular}{|c|c|c|c|c|}\hline
\multirow{3}{*}{\textbf{Dataset}} & \multirow{3}{*}{\textbf{Algorithms}} & \multicolumn{3}{c|}{\textbf{ Error \& Accuracy}} \\
\cline{3-5}
&  & {AbsRel}  $\downarrow$ & {RMSE} $\downarrow$ & $\boldsymbol{\delta < 1.25} $ $\uparrow$  \\ 
\hline 
\multirow{4}{*}{\rotatebox[origin=c]{90}{ \textbf{NYU} \cite{NYU}}} & DPT\cite{DPT}  & 0.110 & 0.392 & 0.864\\

& MultiRes \cite{multires} & 0.102 & 0.347 & 0.921  \\

& ViT-B+ R + AL  & 0.104 & 0.362 & 0.916 \\
 
& DPT-B + R + AL  & \textbf{0.094} & \textbf{0.310} & \textbf{0.945}\\

\hline
\multirow{4}{*}{\rotatebox[origin=c]{90}{ \textbf{KITTI} \cite{Kitti}}} & DPT\cite{DPT}  & 0.114 & 4.773 & 0.849  \\
& MultiRes \cite{multires} & 0.059 & 2.756  & 0.956\\

& ViT-B+ R + AL & 0.078 & 2.933  & 0.915 \\

& DPT-B + R + AL& \textbf{0.056} & \textbf{2.453} & \textbf{0.962}\\
\hline
\multirow{4}{*}{\rotatebox[origin=c]{90}{ \textbf{HRSD}}}& DPT\cite{DPT}  & 0.127 & 0.494  & 0.822  \\

& MultiRes \cite{multires} & 0.096 &  0.339  & 0.920 \\

& ViT-B + R + AL & 0.099 & 0.322 & 0.912 \\
 
& DPT-B + R + AL & \textbf{0.074} & \textbf{0.288}  & \textbf{0.921} \\

\hline
\end{tabular}
\caption{ Quantitative comparison on three RGB-D datasets. Here DPT\cite{DPT} and MultiRes \cite{multires} results are obtained using the author's weights. ViT-B + R + AL and DPT-B + R + AL are the variants of ViT\cite{ViT} and DPT\cite{DPT} trained on the proposed HRSD datasets.}
\label{outcomp}
\end{table}
\par
\begin{figure*}[h!]
\centering
\begin{subfigure}{.16\textwidth}
  \includegraphics[width=\linewidth,height=28.5mm]{images/comparision/new-hm/NYUV244.jpg}  \\
   \includegraphics[width=\linewidth,height=28.5mm]{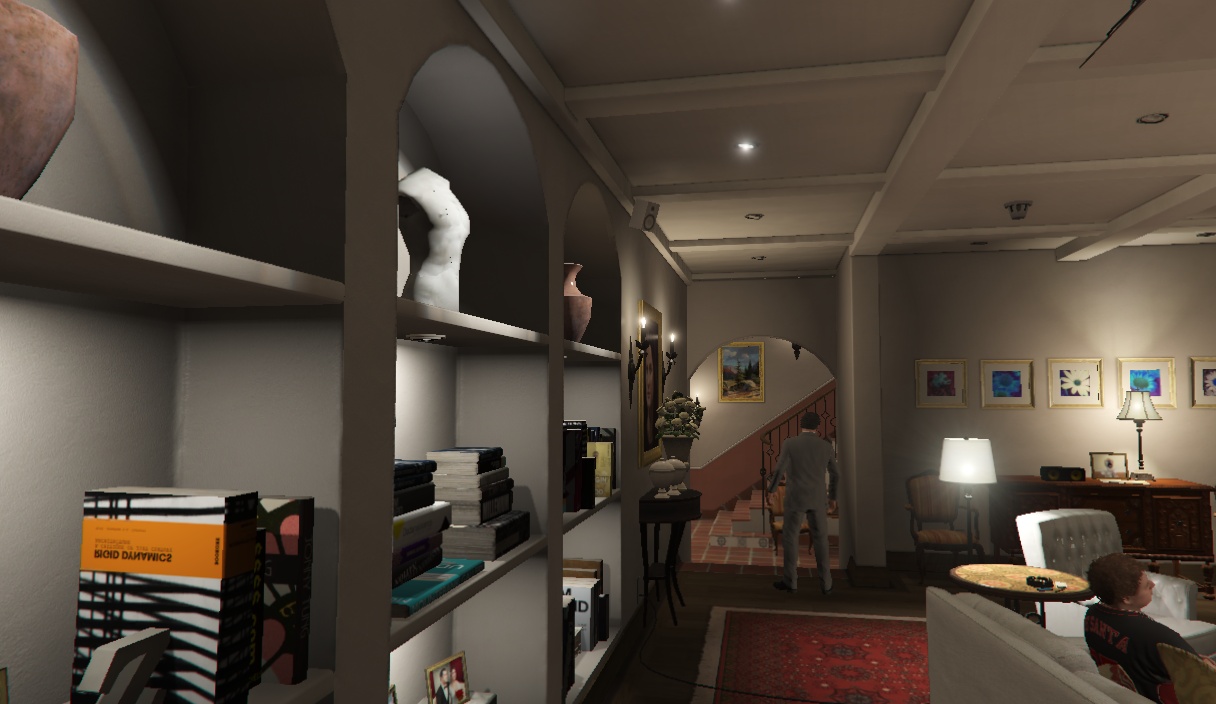} \\
   \includegraphics[width=0.98\linewidth,height=28.5mm]
    {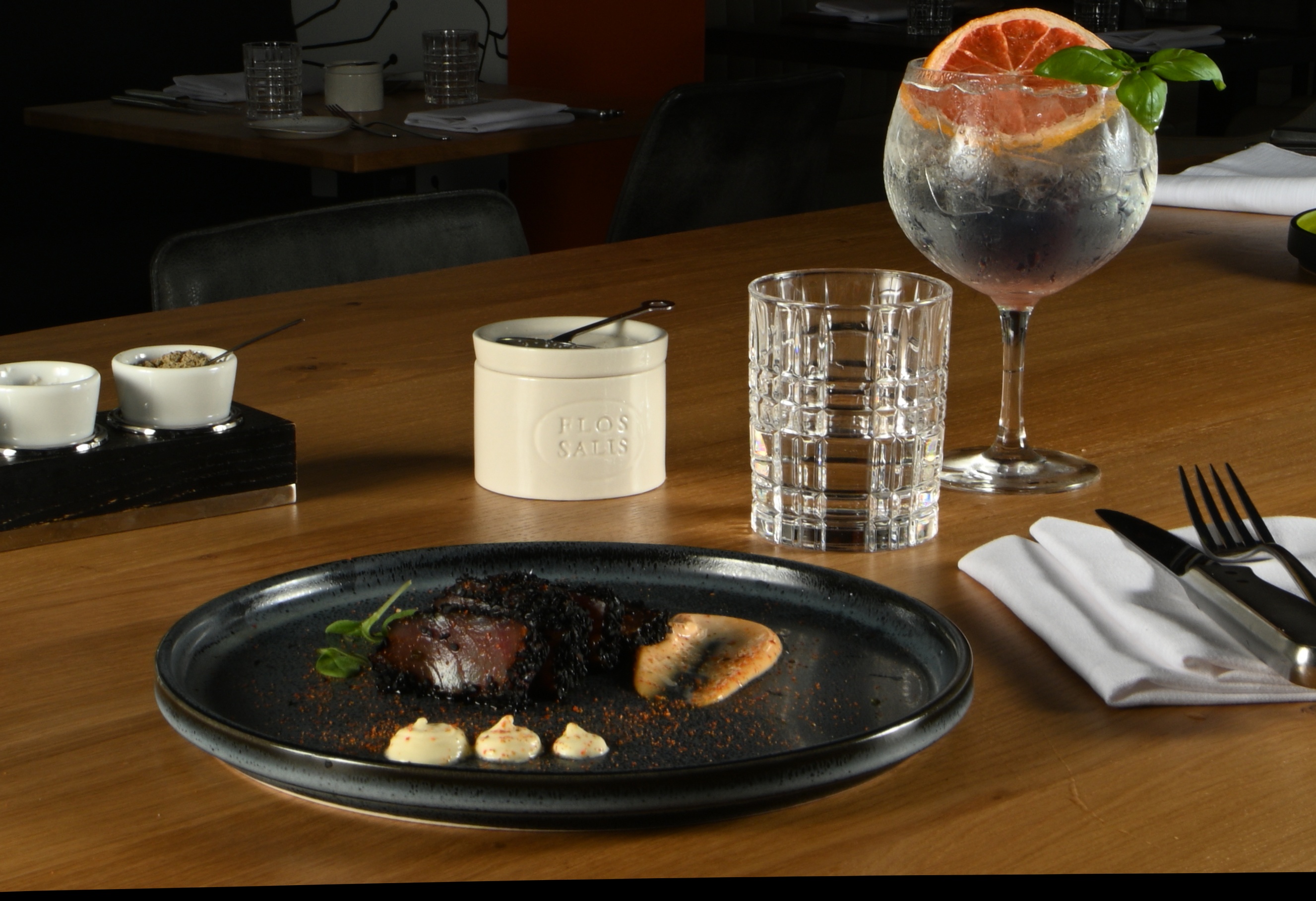}
   
   \caption{RGB}
\end{subfigure}
 \begin{subfigure}{.16\textwidth}
    \includegraphics[width=0.98\linewidth,height=28.5mm]{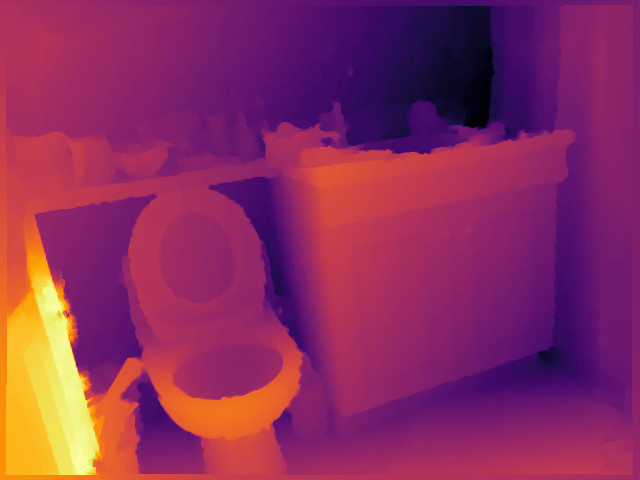}\\
    \includegraphics[width=0.98\linewidth,height=28.5mm]
    {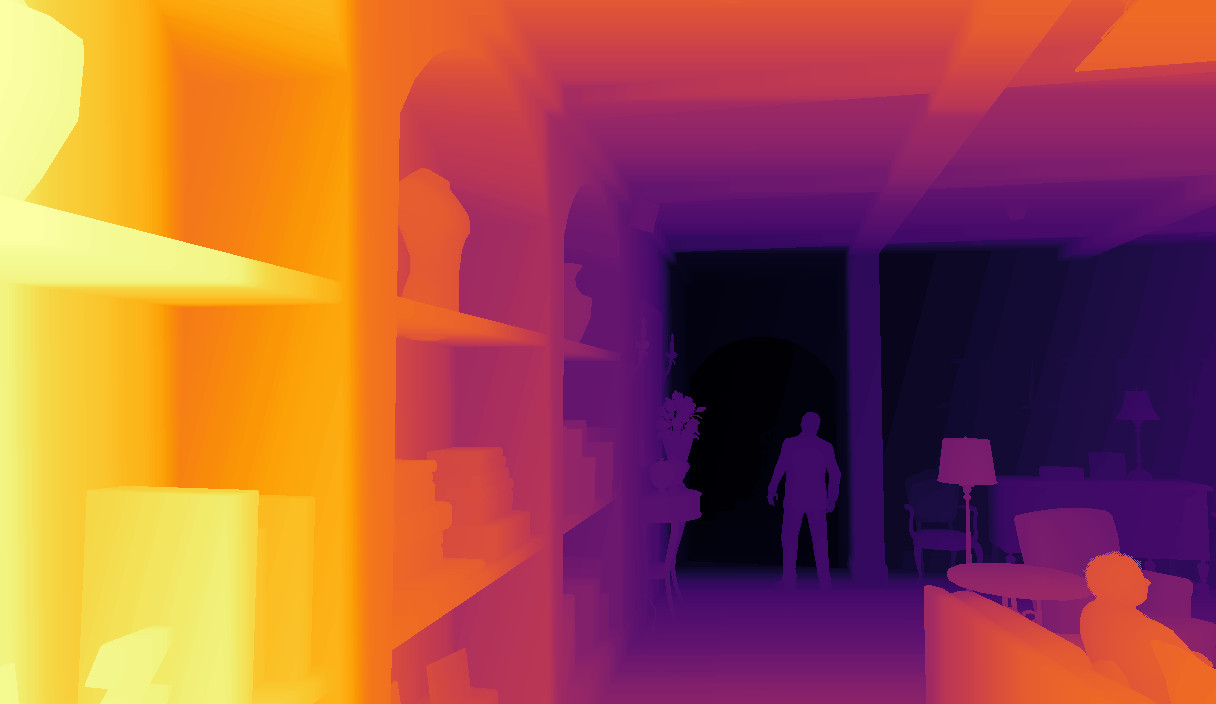} \\    
    \includegraphics[width=0.98\linewidth,height=28.5mm]
    {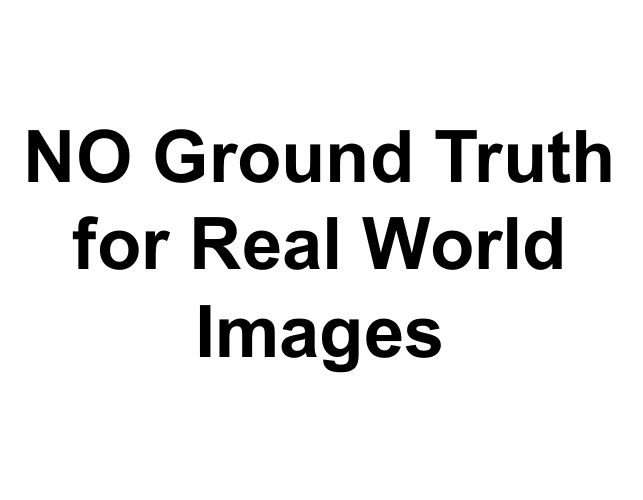}
    \caption{GT-Depth}
\end{subfigure}
\begin{subfigure}{.16\textwidth}
    \includegraphics[width=0.98\linewidth,height=28.5mm]{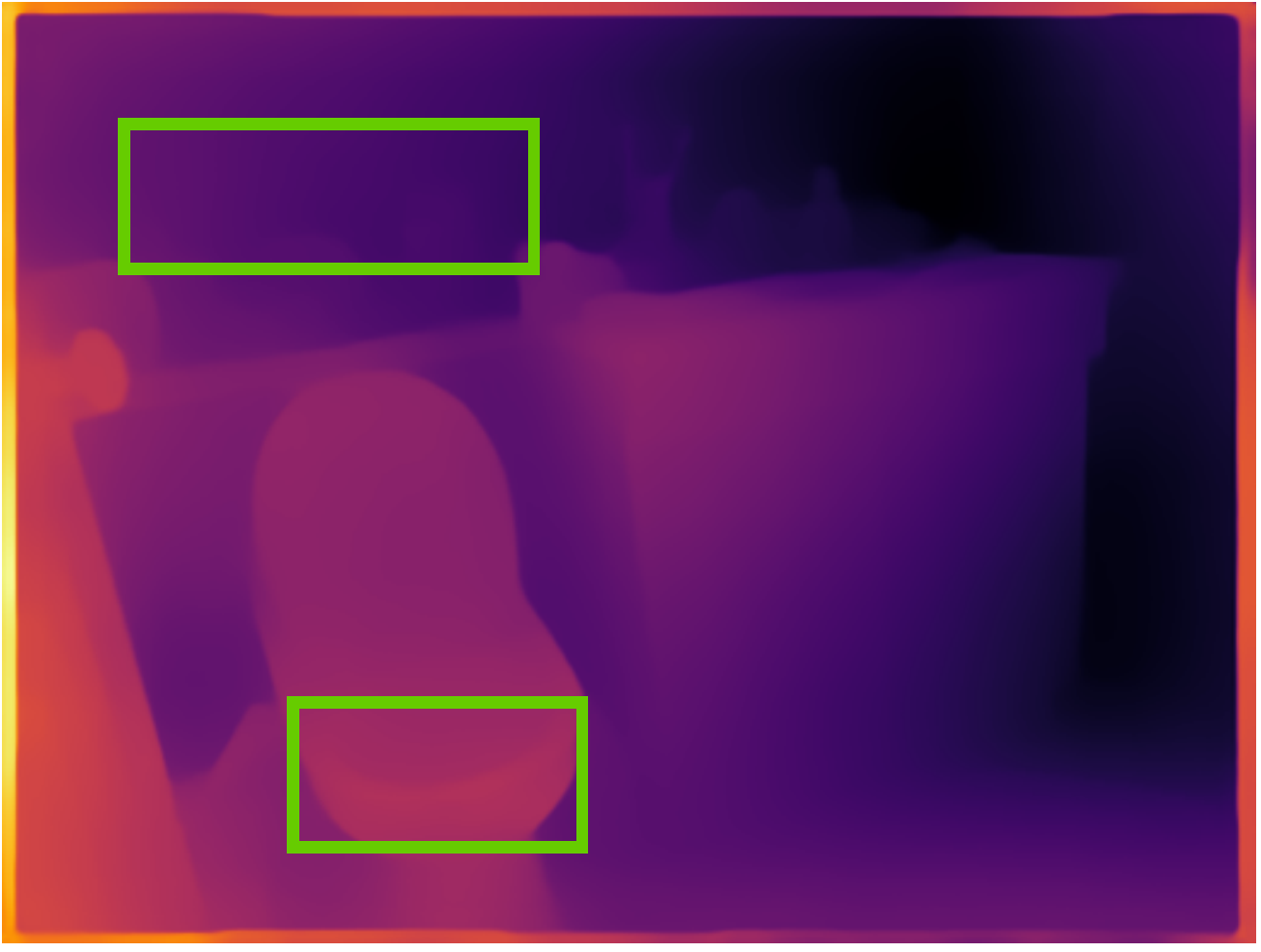} \\
    \includegraphics[width=0.98\linewidth,height=28.5mm]{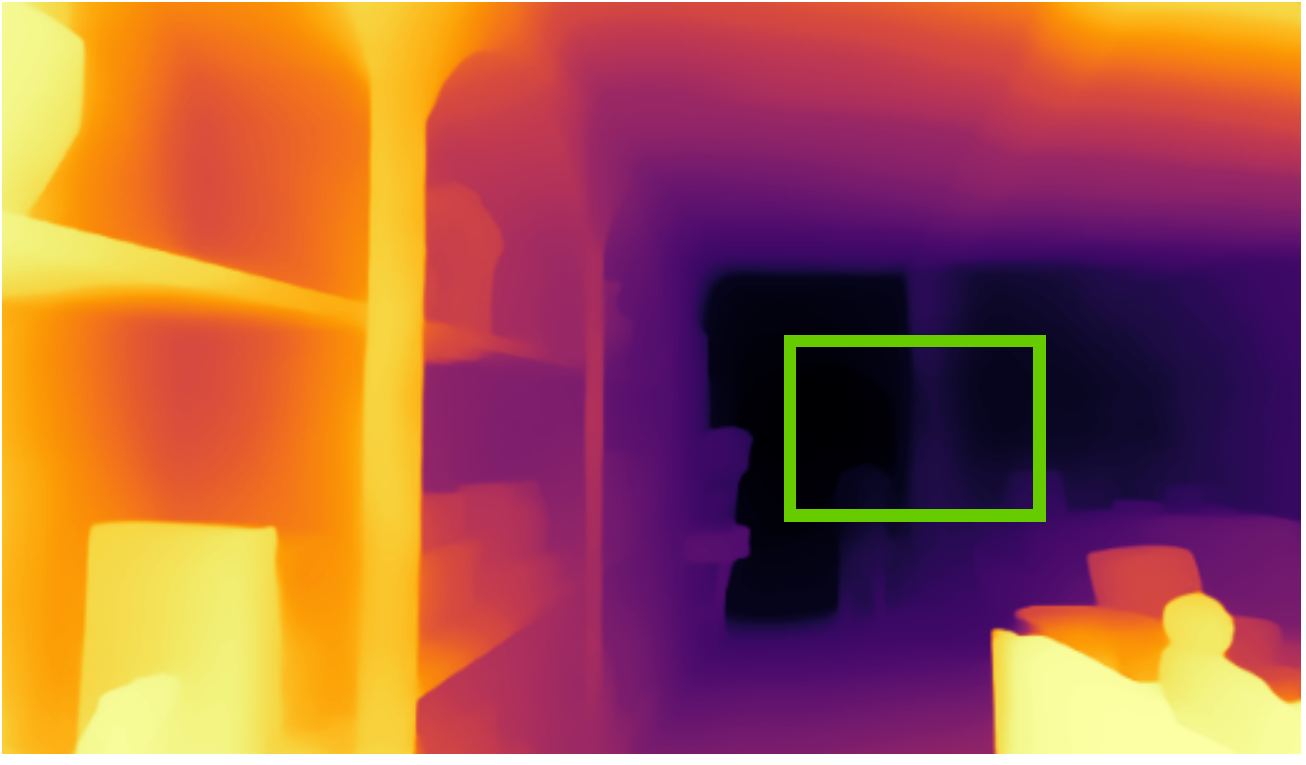}\\
     \includegraphics[width=0.98\linewidth,height=28.5mm]{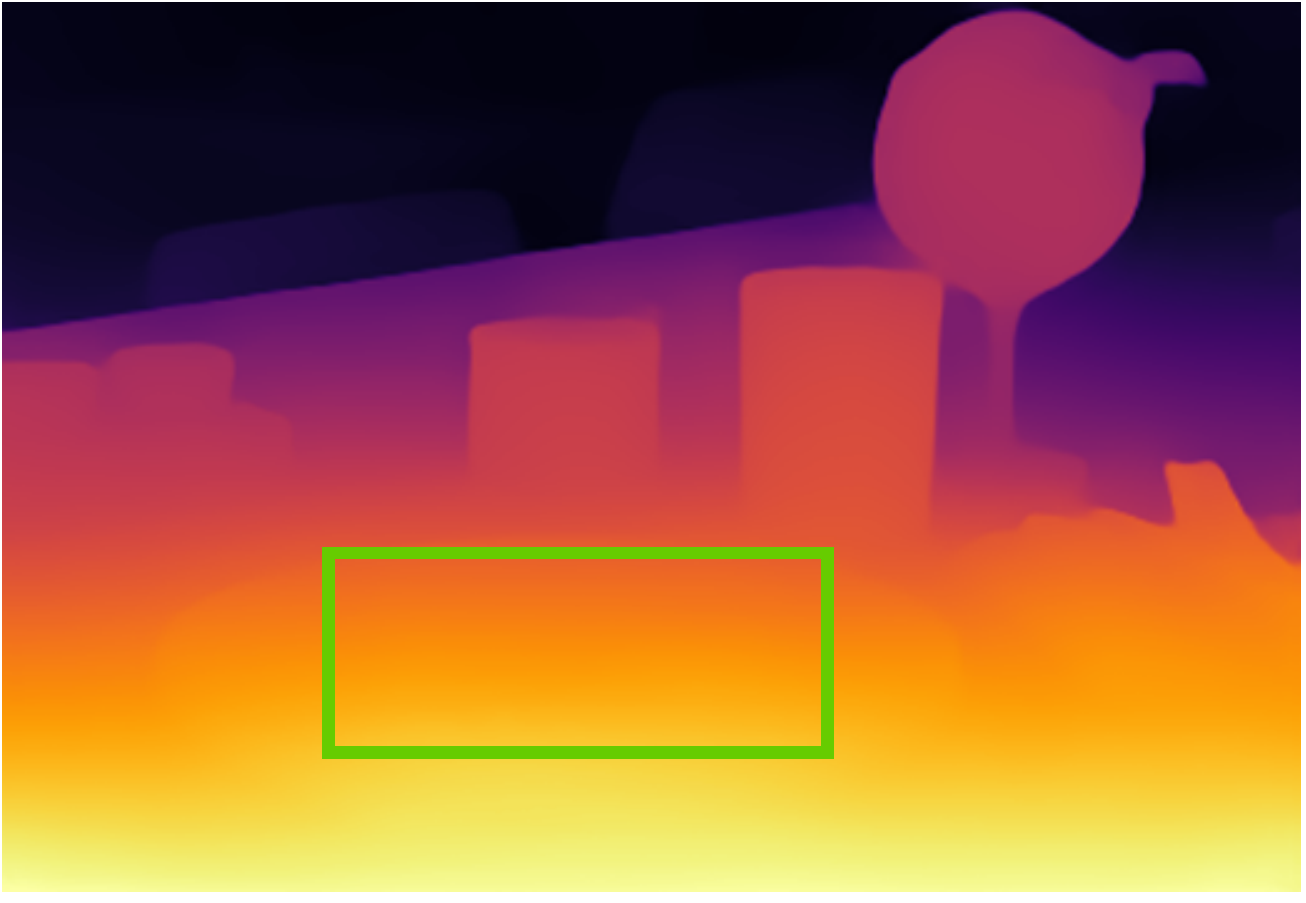}
    
    \caption{DPT \cite{DPT}}
\end{subfigure}
\begin{subfigure}{.16\textwidth}
   \includegraphics[width=0.98\linewidth,height=28.5mm]{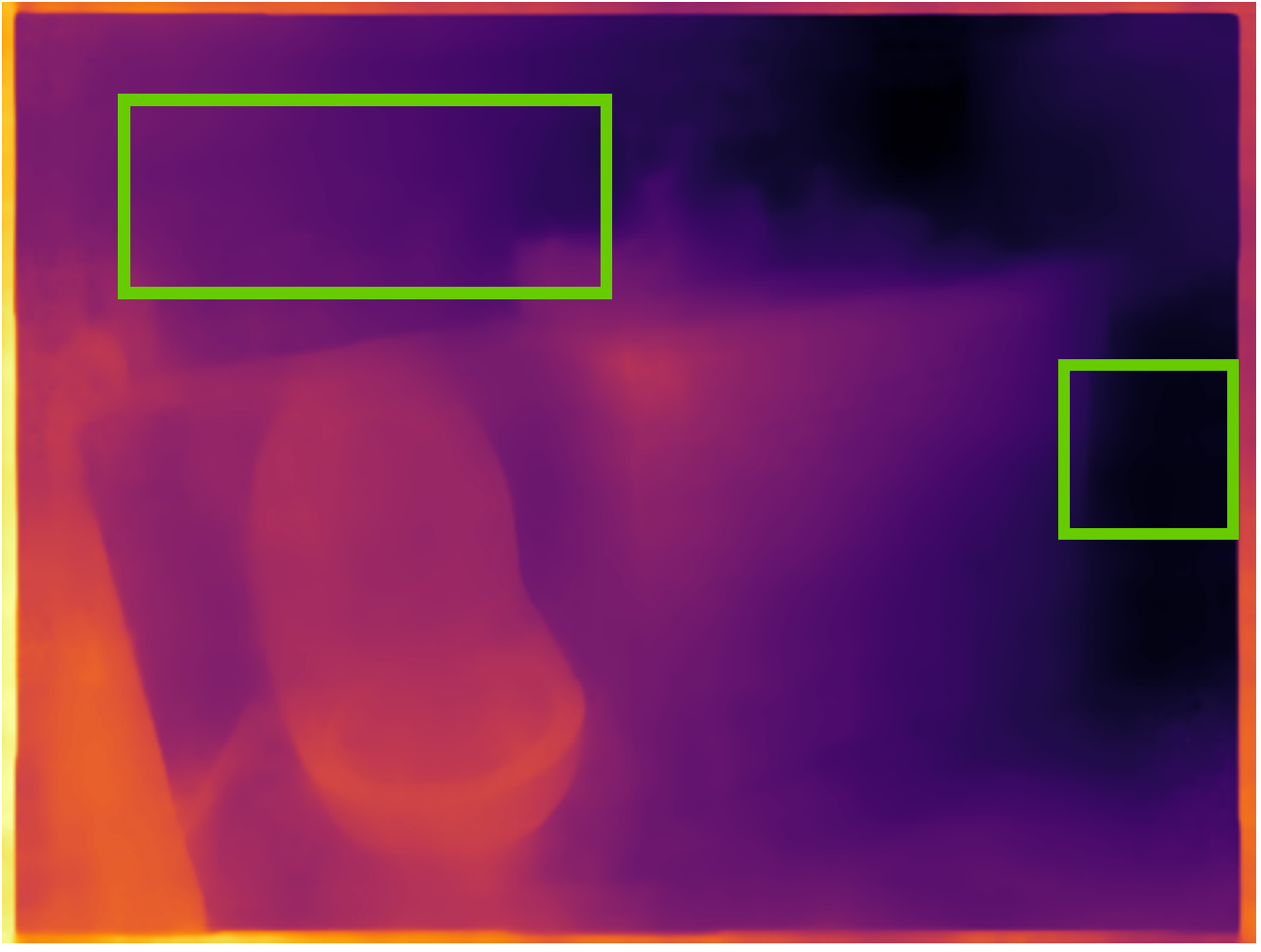} \\
   \includegraphics[width=0.98\linewidth,height=28.5mm]{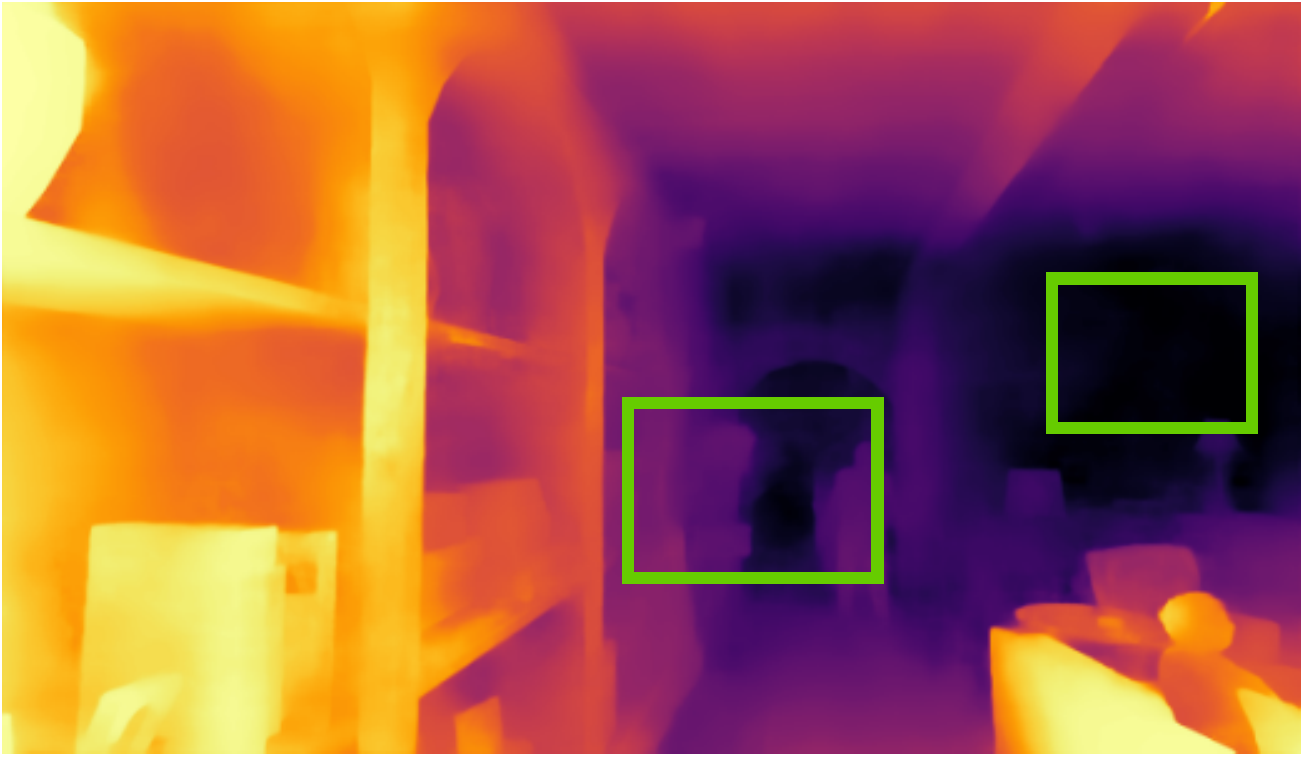}\\
   \includegraphics[width=0.98\linewidth,height=28.5mm]{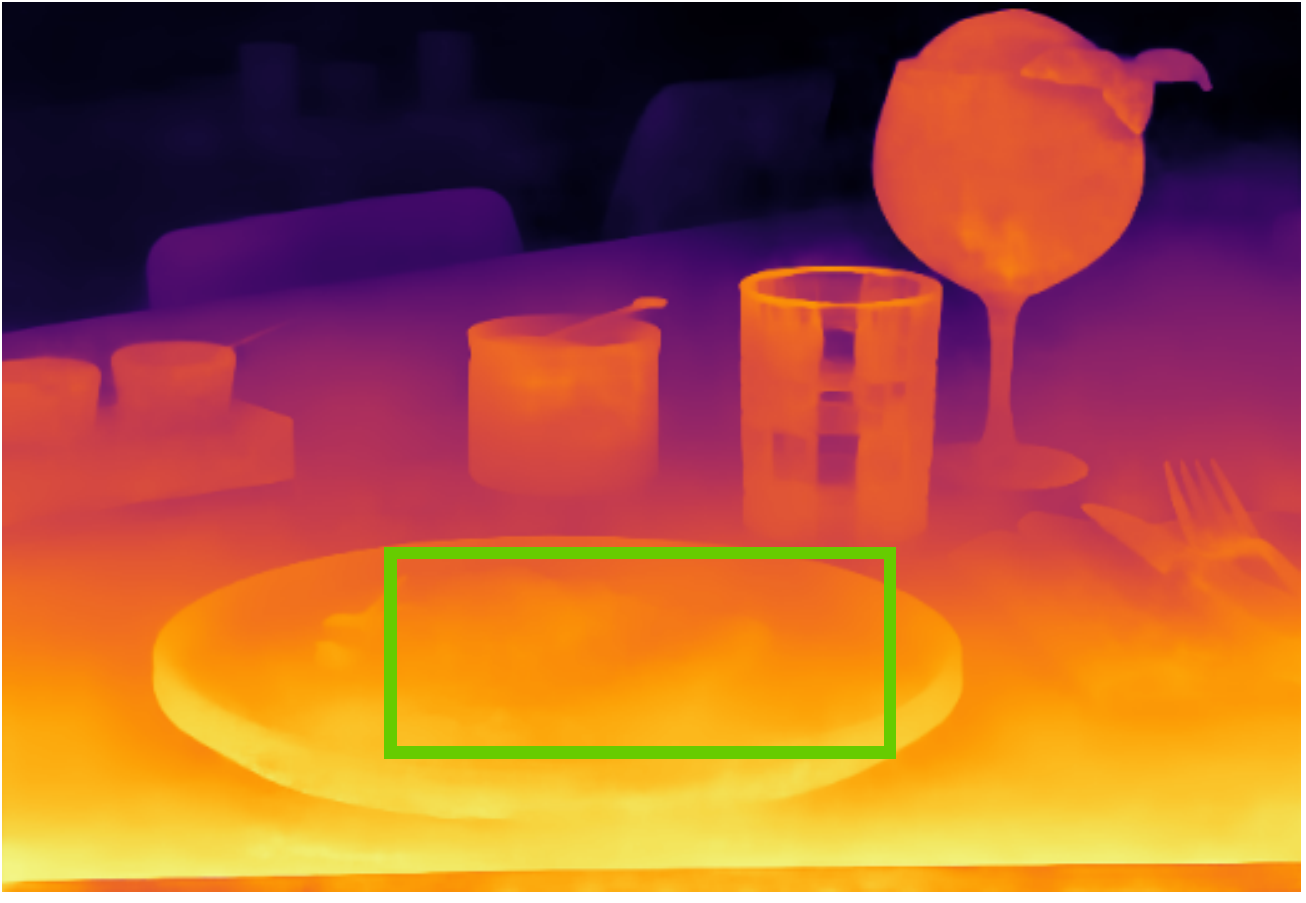}
   
   \caption{MultiRes \cite{multires}}
\end{subfigure}
\begin{subfigure}{.16\textwidth}
   \includegraphics[width=0.98\linewidth,height=28.5mm]{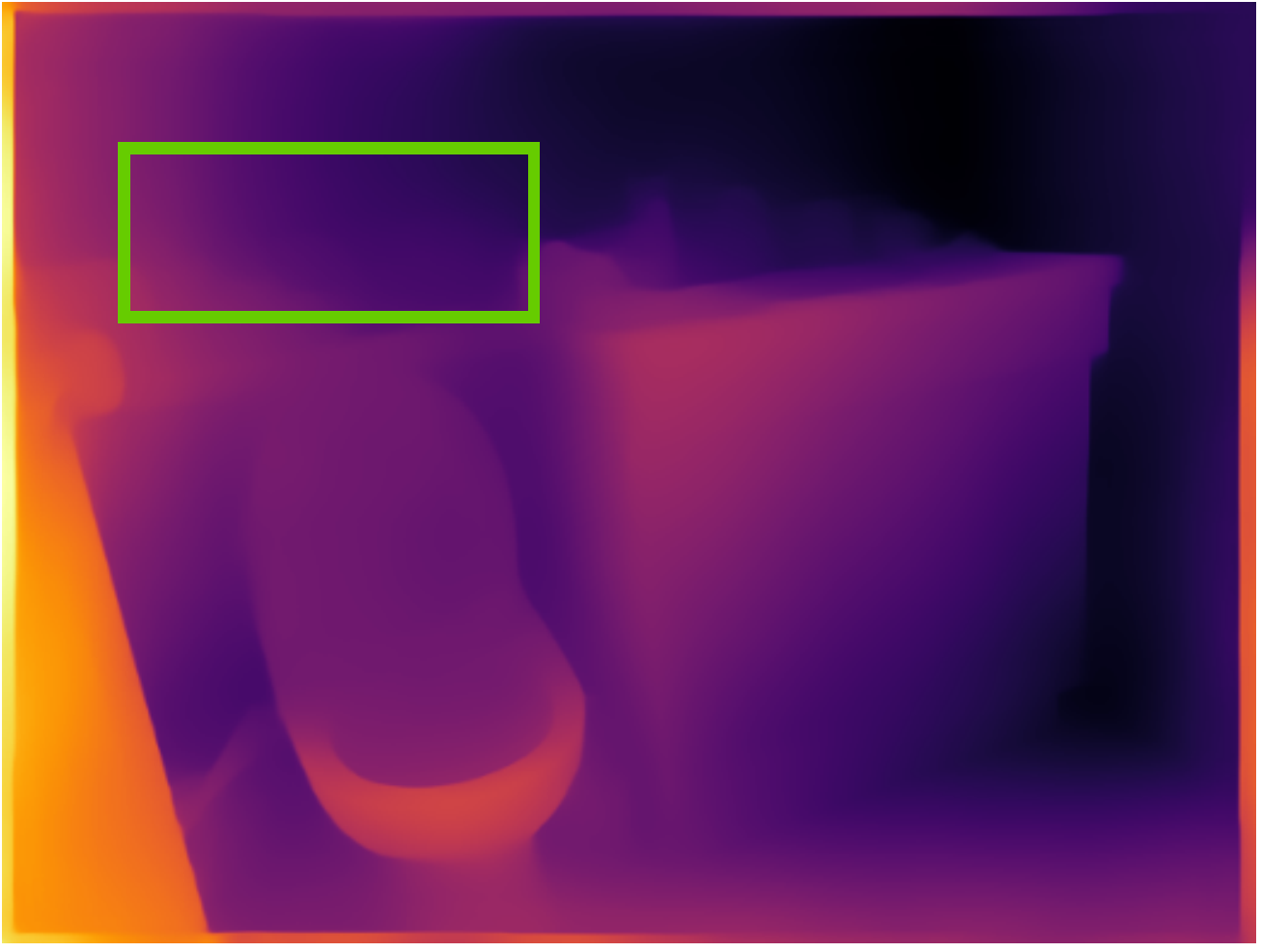} \\
   \includegraphics[width=0.98\linewidth,height=28.5mm]{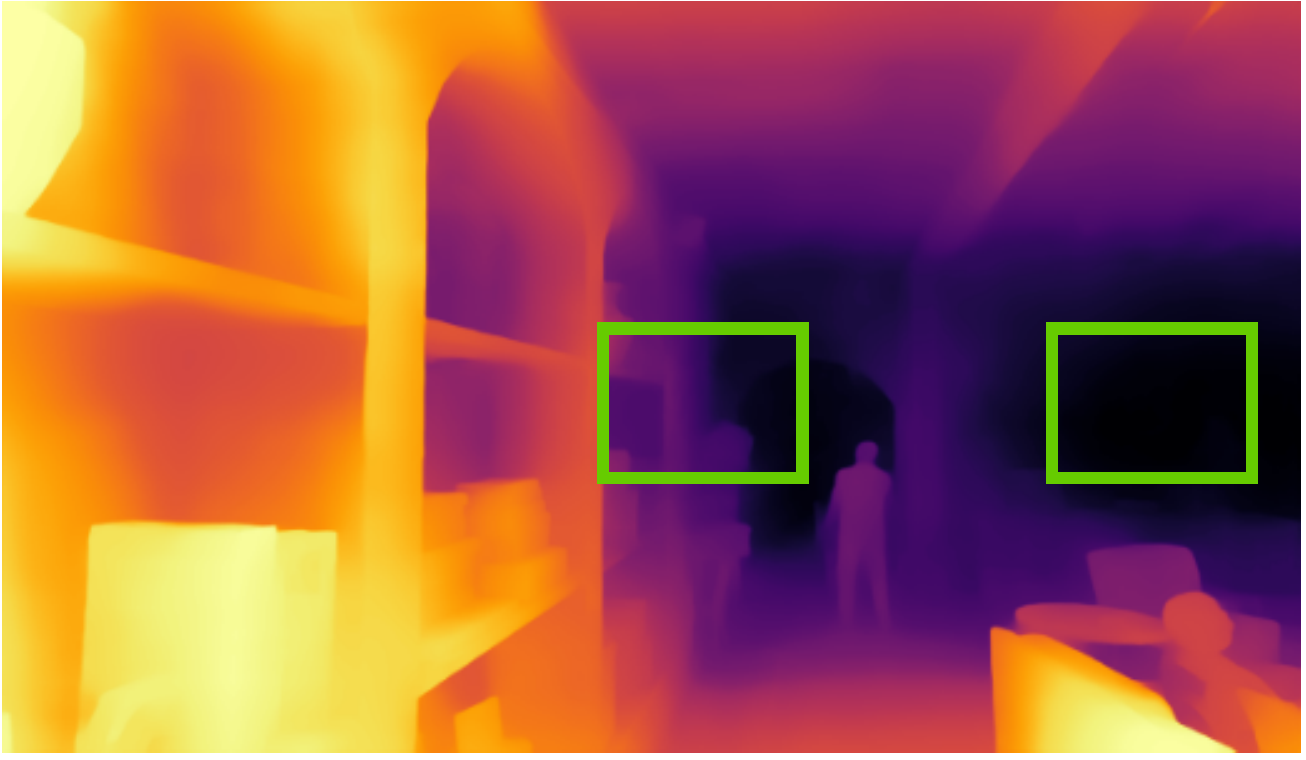}\\
   \includegraphics[width=0.98\linewidth,height=28.5mm]{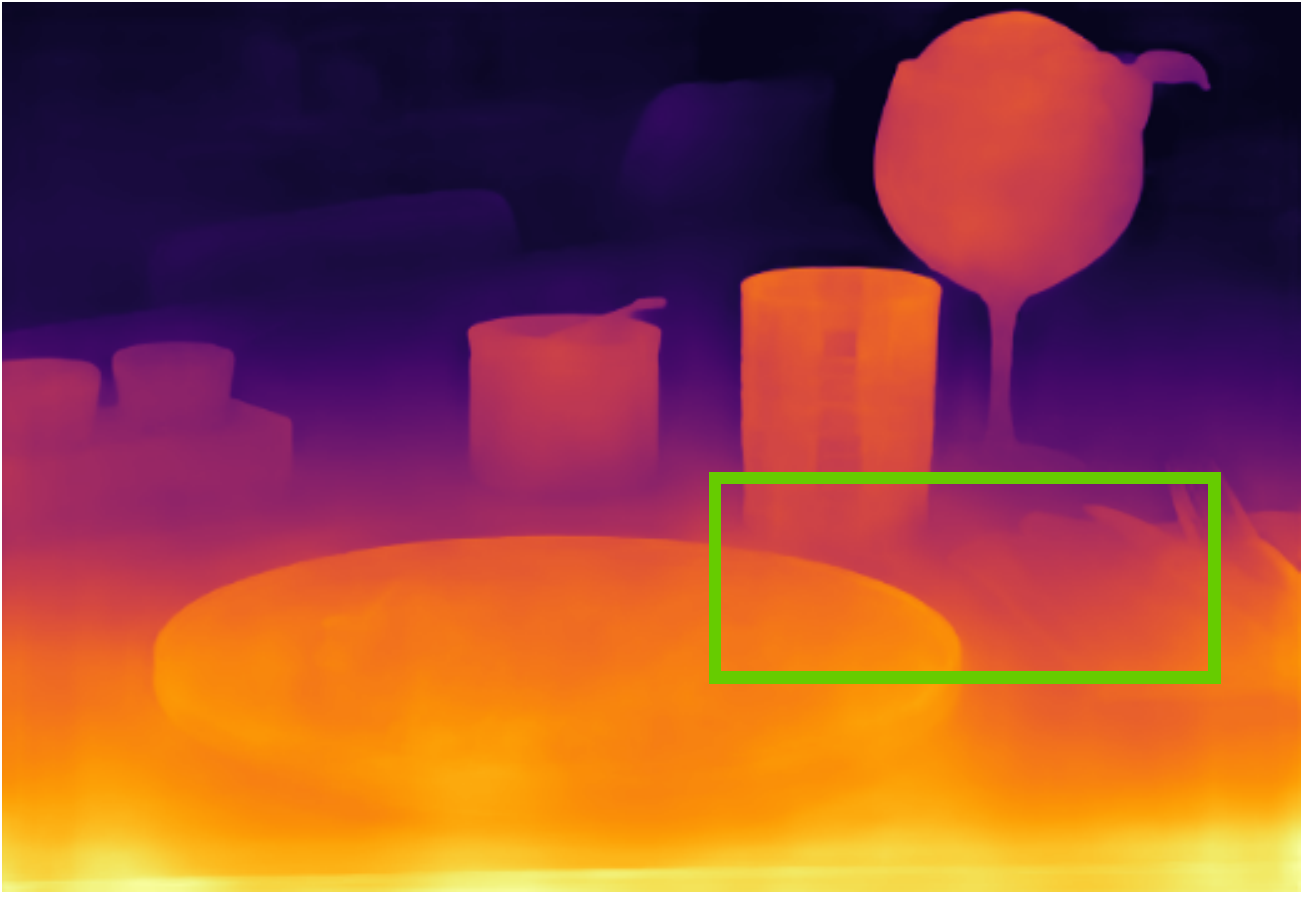}   \caption{ViT-B + R + AL}
\end{subfigure}
\begin{subfigure}{.16\textwidth}
   \includegraphics[width=0.98\linewidth,height=28.5mm]{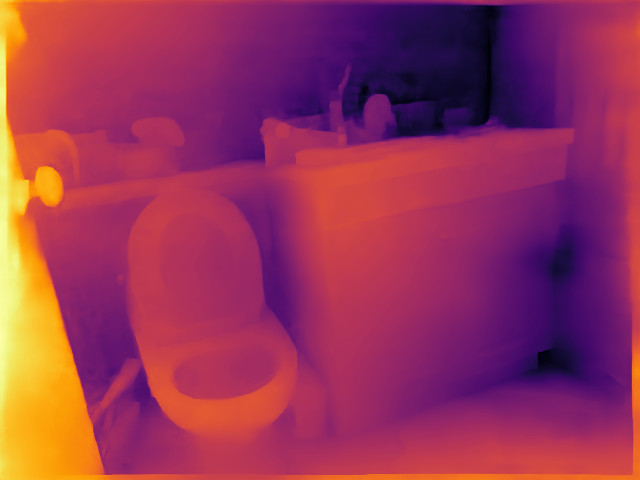} \\ 
   \includegraphics[width=0.98\linewidth,height=28.5mm]{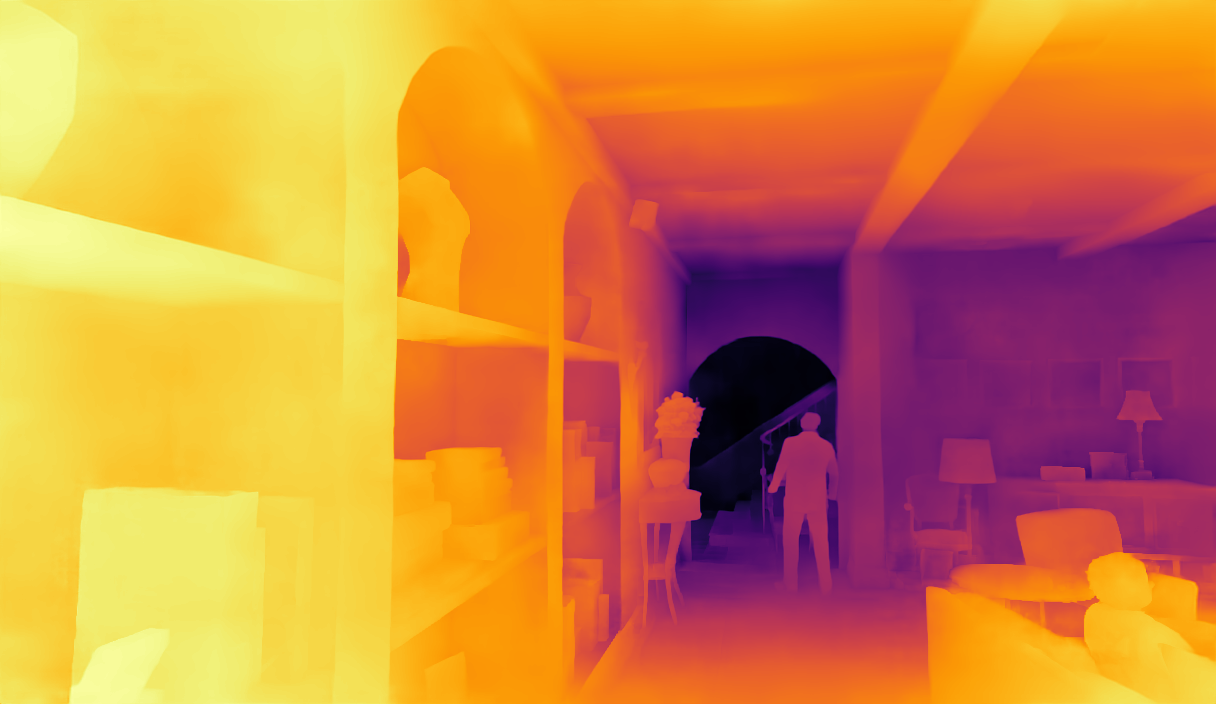}\\
   \includegraphics[width=0.98\linewidth,height=28.5mm]{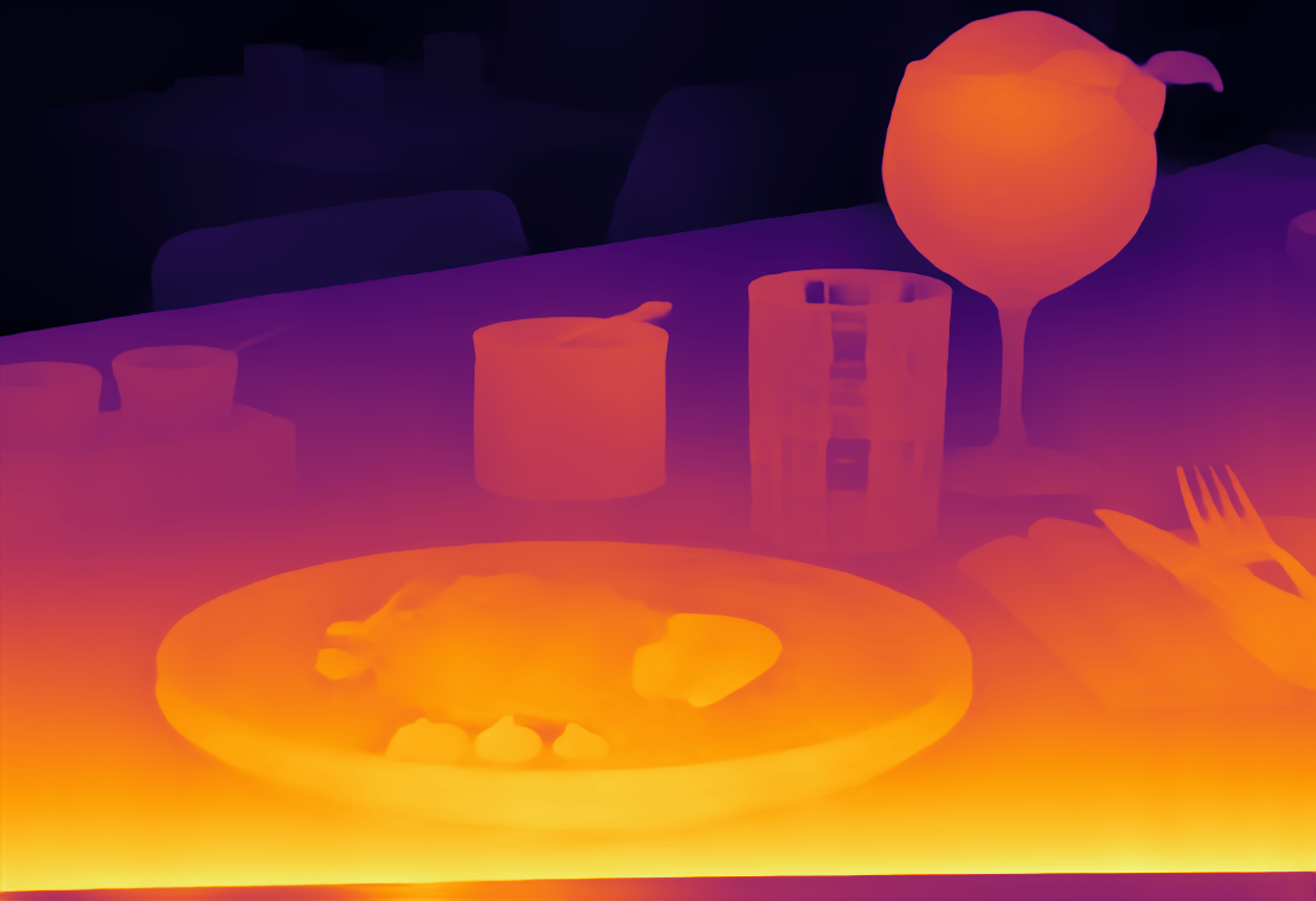}
   \caption{DPT-B + R + AL}
\end{subfigure}
\caption{Indoor Scenes. $1^{\text {st}}$Row:- NYU\cite{NYU}. $2^{\text {nd}}$Row:- HRSD indoor. $3^{\text {rd}}$ Row:- RealWorld. Our DPT-B + R + AL gives a consistent depth map across all regions and displays sharp structure for overall objects i.e. items on the table in the real-world images. Original DPT \cite{DPT} fails to identify objects in the background as shown by the green rectangles i.e. no structure of human in HRSD indoor. Multires \cite{multires} leads to an inconsistent depth map highlighted by green rectangles i.e. the toilet seat in NYU image.}
\label{interimg}
\end{figure*}

% \indoor[table\indoor_comarison]
\vspace{.5cm}
\subsection{Evaluation}
Like \cite{eigen,DPT,midas}, we use three error metrics, such as RMSE, AbsRel, and percentage of correct pixels, to evaluate the performance of depth maps. 

\textbf{Datasets for evaluation}
To demonstrate the competitiveness of our approach, we evaluate the methods against KITTI \cite{Kitti} and  NYU V2 \cite{NYU} RGB-D datasets. These are the standard datasets for outdoor and indoor scenes, respectively. KITTI dataset contains 697 images ($1216 \times 352$) with re-projected Lidar points as sparse depth maps and NYU V2 contains 694 images ($640 \times 480$). In addition, we evaluate depth maps on 10,000 images ($1920 \times 1080$) of our HRSD dataset as high-resolution datasets like \cite{midas,diode} are unavailable to the public.

\subsection{Results}
\par
\textbf{Quantitative Results}
To show the effect of the proposed HRSD datasets on different variants of training, we make a quantitative comparison of the evaluation of different datasets and present them in Table \ref{intercomp} and Table \ref{outcomp}. In Table \ref{intercomp}, we compare six variants of training performed using DPT \cite{DPT} encoder weights \& ViT \cite{ViT} weights as described earlier. The addition of the feature module and attention-loss performs better in all three datasets in all metrics. In Table \ref{outcomp}, we compare the original DPT \cite{DPT} and MultiRes \cite{multires} with the proposed variant of ViT (ViT-B+ R + AL) and DPT (DPT-B+ R + AL). From Table \ref{outcomp}, we can conclude that the proposed variants are better in almost all the datasets. This indicates the effectiveness of the proposed HRSD datasets, which results in a lower absolute error and higher accuracy.

\begin{figure*}[h!]
\centering
\begin{subfigure}{.16\textwidth}
  \includegraphics[width=\linewidth,height=28.5mm]{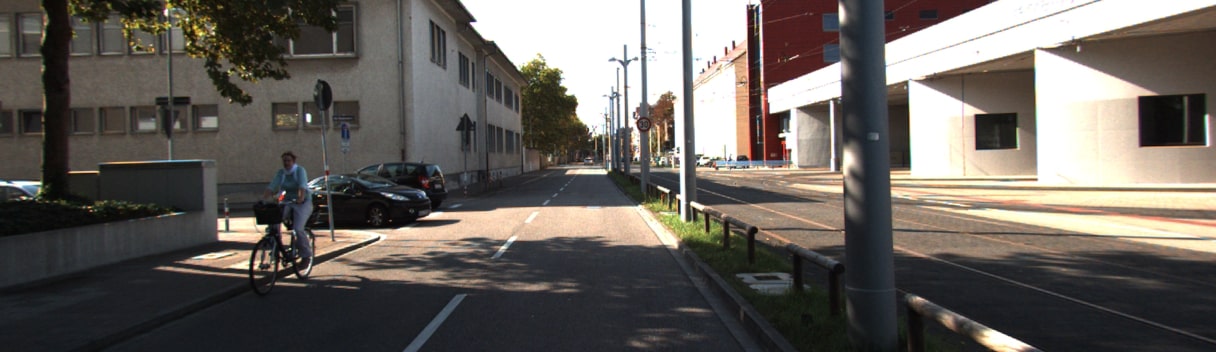}  \\
   \includegraphics[width=\linewidth,height=28.5mm]
   {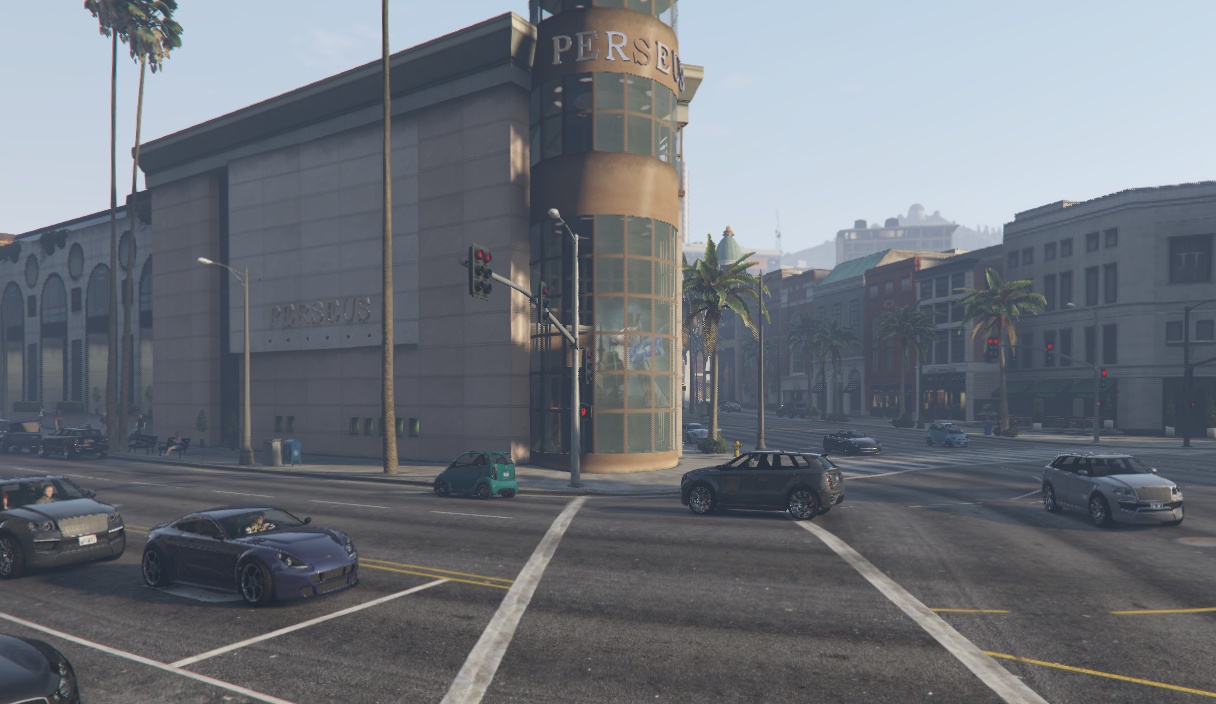}  \\
   \includegraphics[width=\linewidth,height=28.5mm]{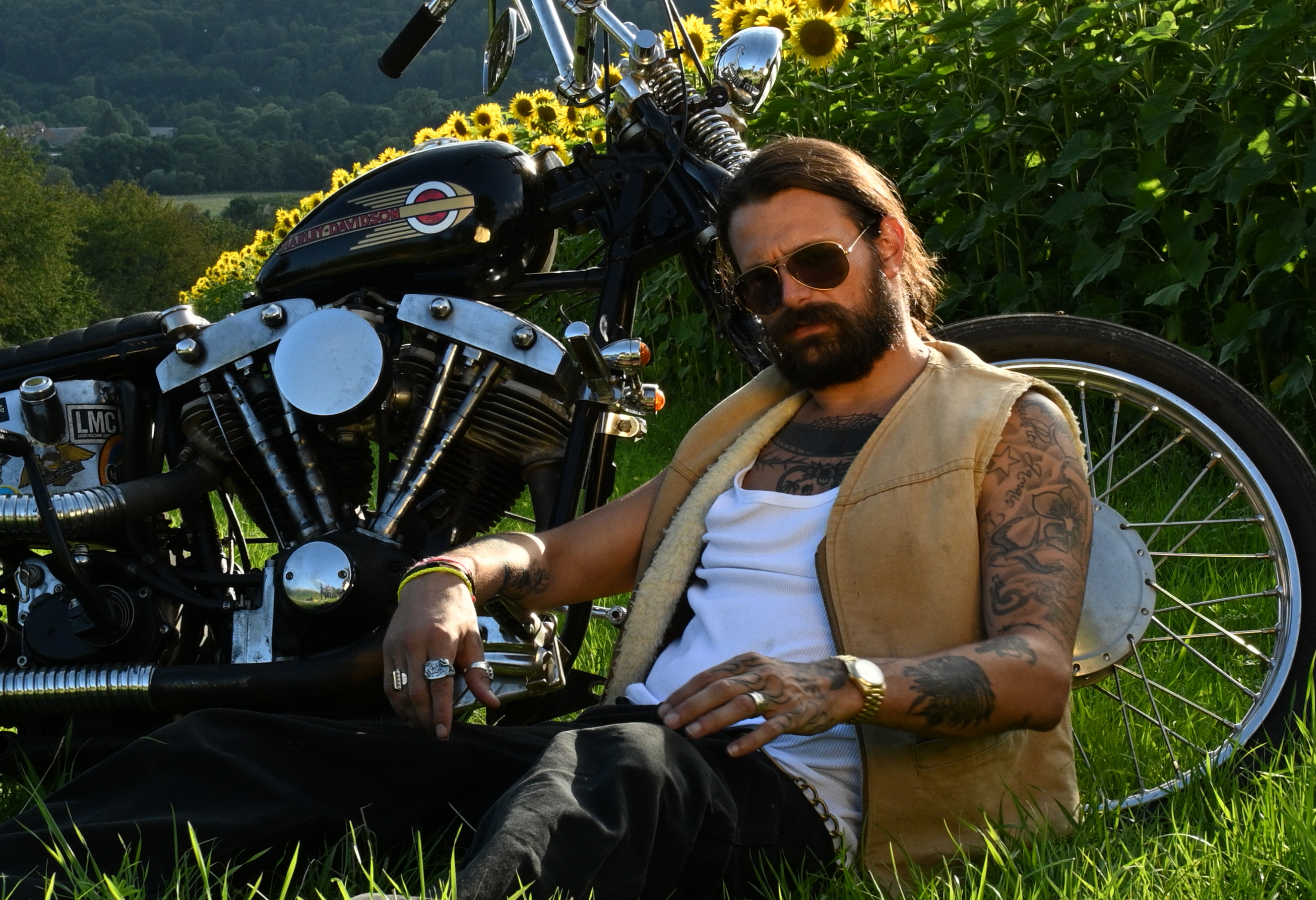}

   \caption{RGB}
\end{subfigure}
\begin{subfigure}{.16\textwidth}
    \includegraphics[width=0.98\linewidth,height=28.5mm]{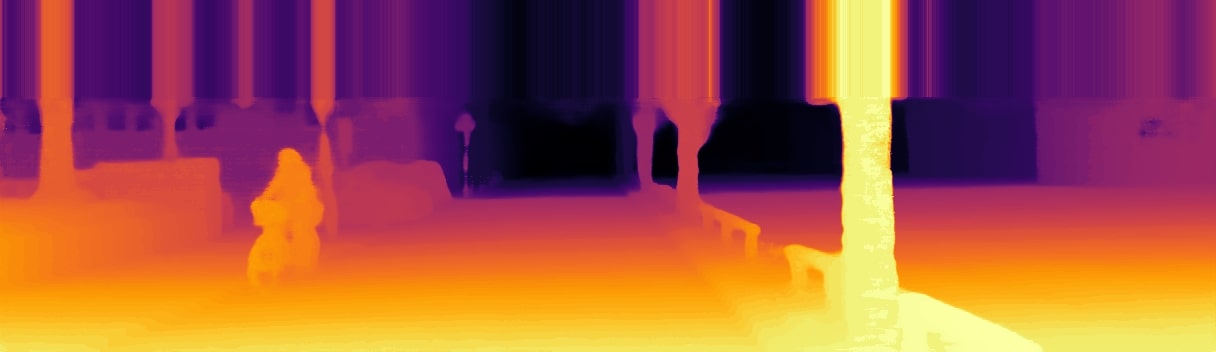} \\
    \includegraphics[width=0.98\linewidth,height=28.5mm]{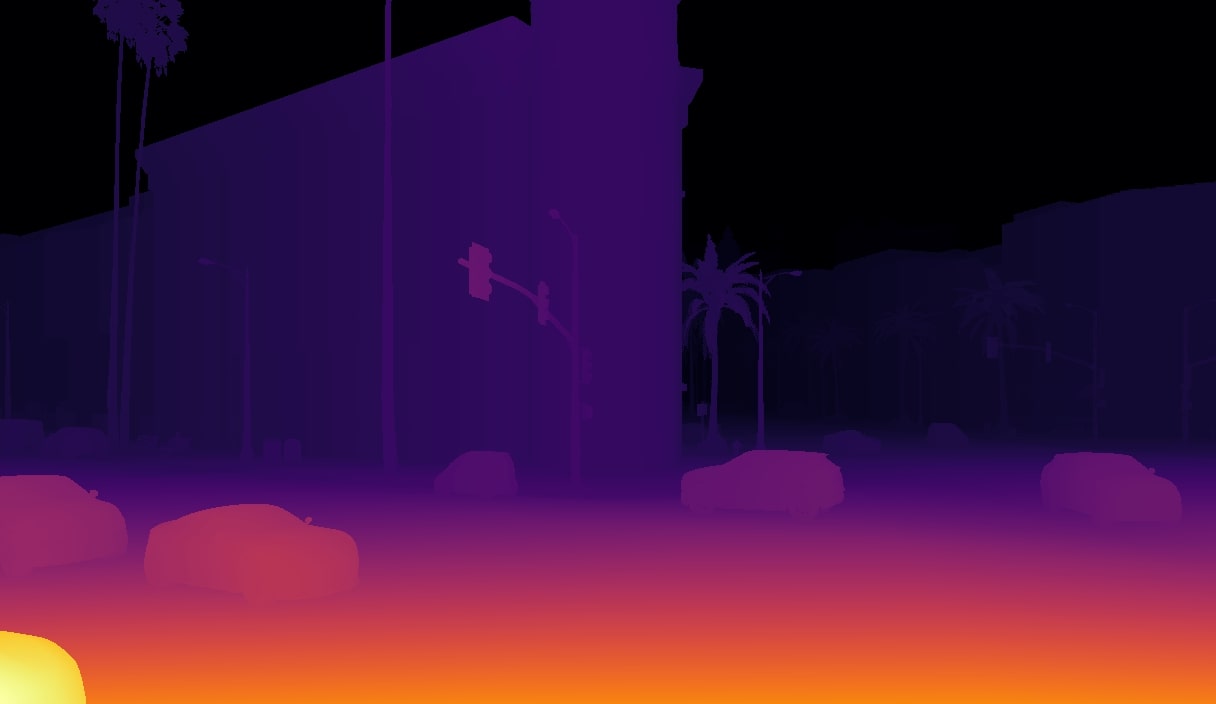}  \\
   \includegraphics[width=\linewidth,height=28.5mm]{images/empty.pdf}
    \caption{GT-Depth}
\end{subfigure}
\begin{subfigure}{.16\textwidth}
   \includegraphics[width=0.98\linewidth,height=28.5mm]{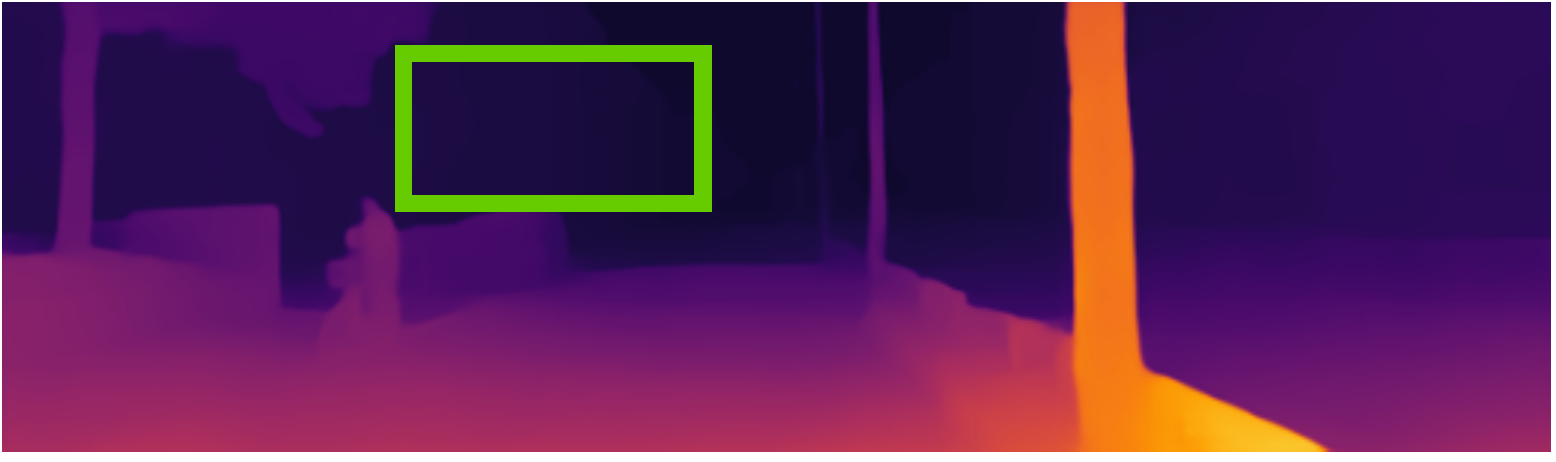} \\
   \includegraphics[width=0.98\linewidth,height=28.5mm]{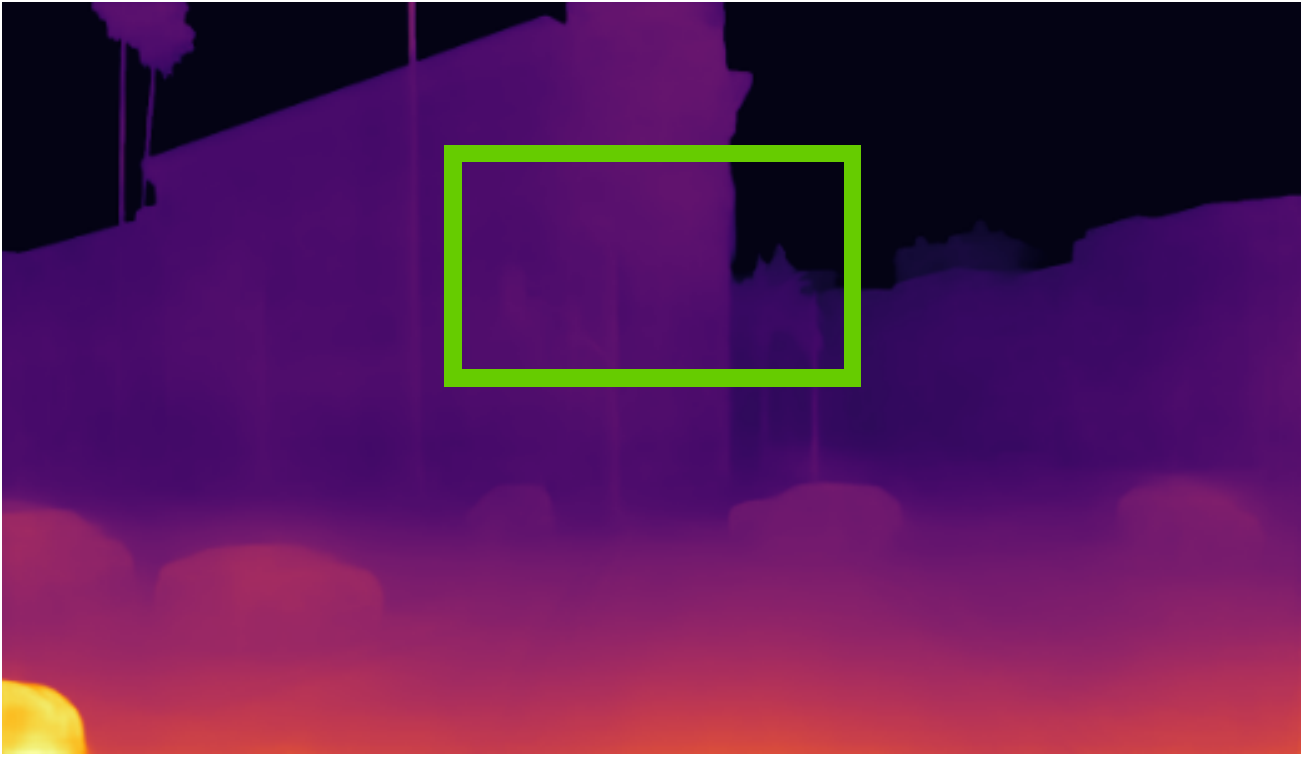}  \\
   \includegraphics[width=\linewidth,height=28.5mm]{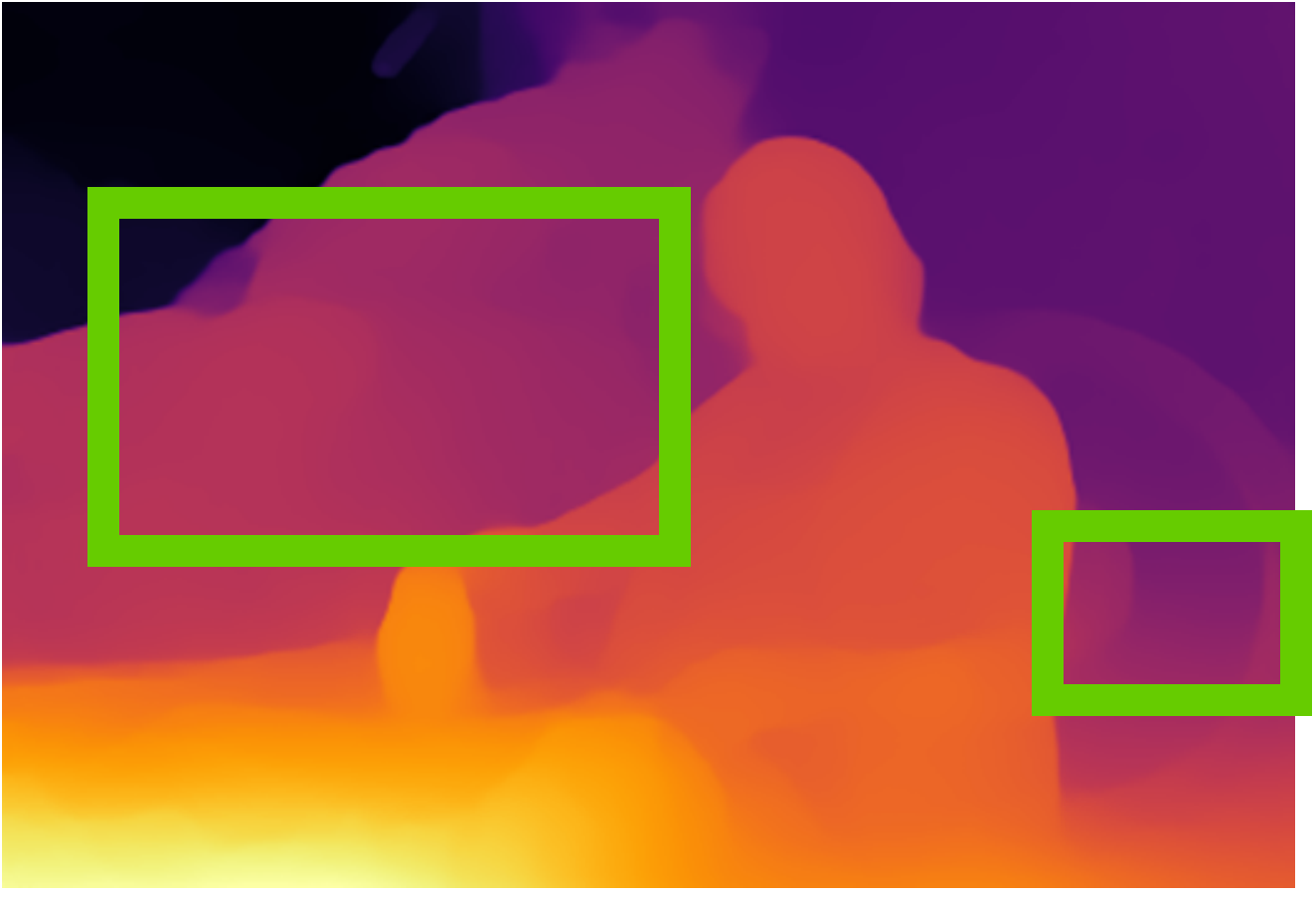}
   \caption{DPT \cite{DPT}}
\end{subfigure}
\begin{subfigure}{.16\textwidth}
   \includegraphics[width=0.98\linewidth,height=28.5mm]{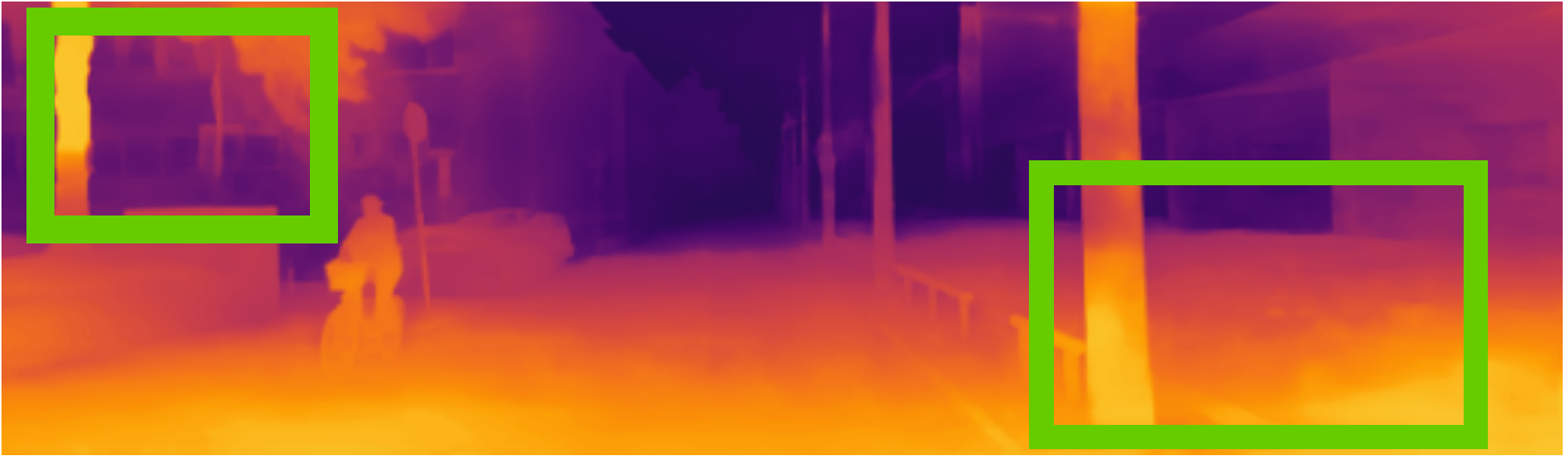} \\
   \includegraphics[width=0.98\linewidth,height=28.5mm]{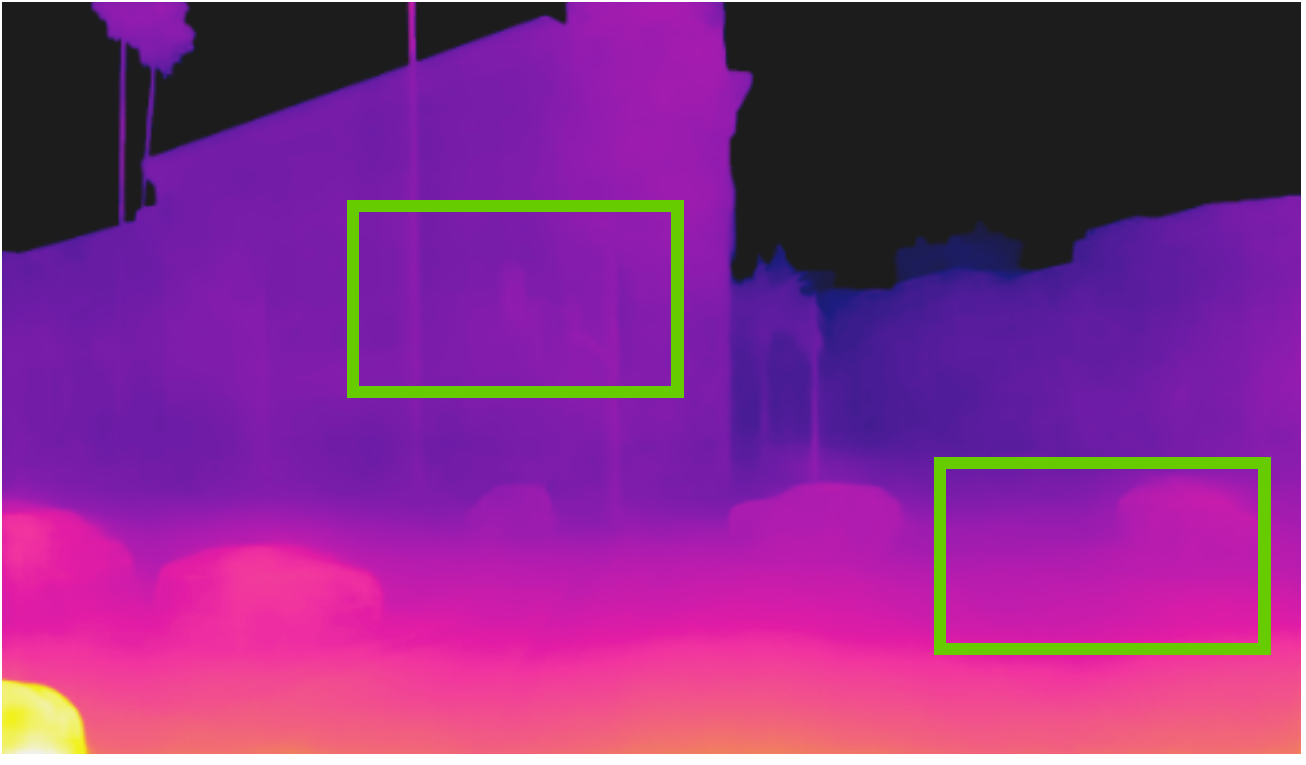}  \\
   \includegraphics[width=\linewidth,height=28.5mm]{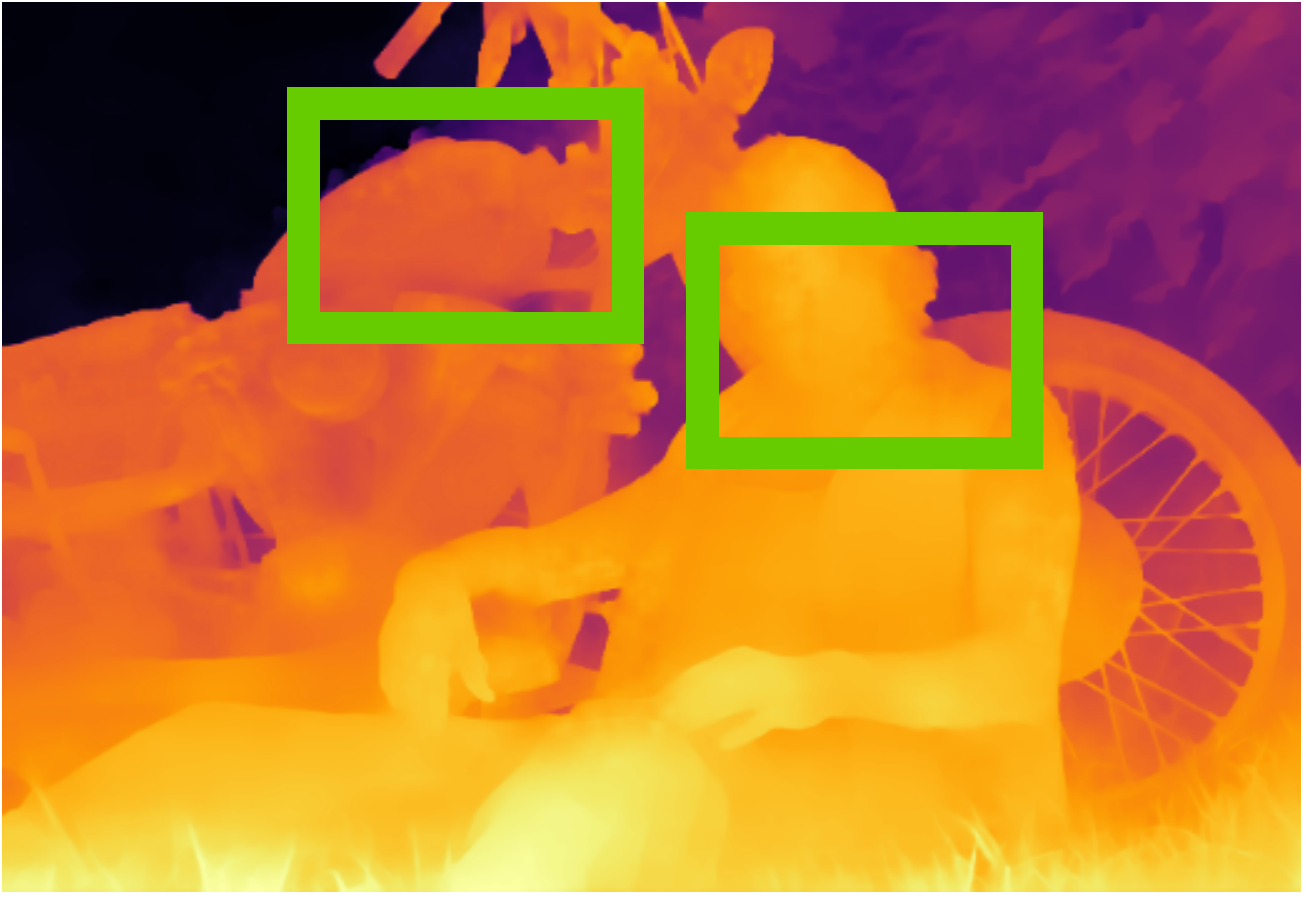}
   \caption{MultiRes \cite{multires}}
\end{subfigure}
\begin{subfigure}{.16\textwidth}
   \includegraphics[width=0.98\linewidth,height=28.5mm]{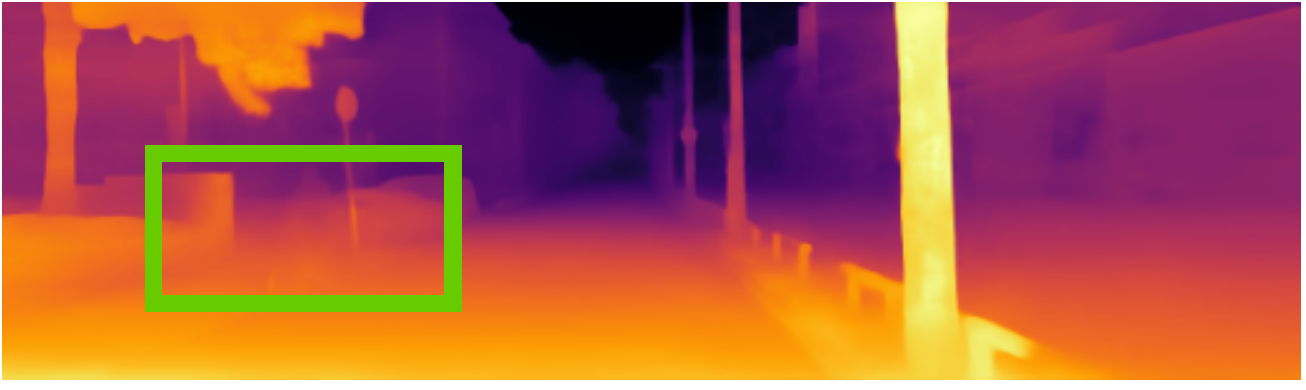} \\
   \includegraphics[width=0.98\linewidth,height=28.5mm]{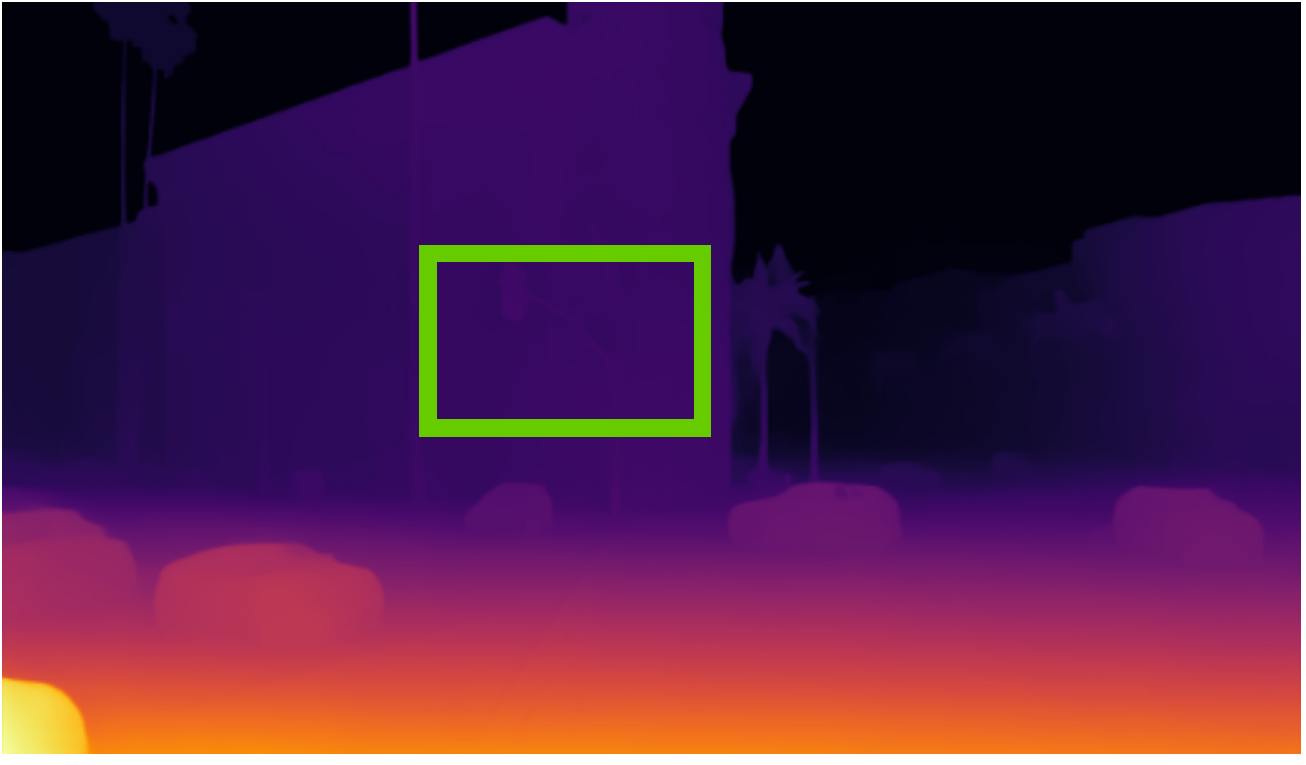}  \\
   \includegraphics[width=\linewidth,height=28.5mm]{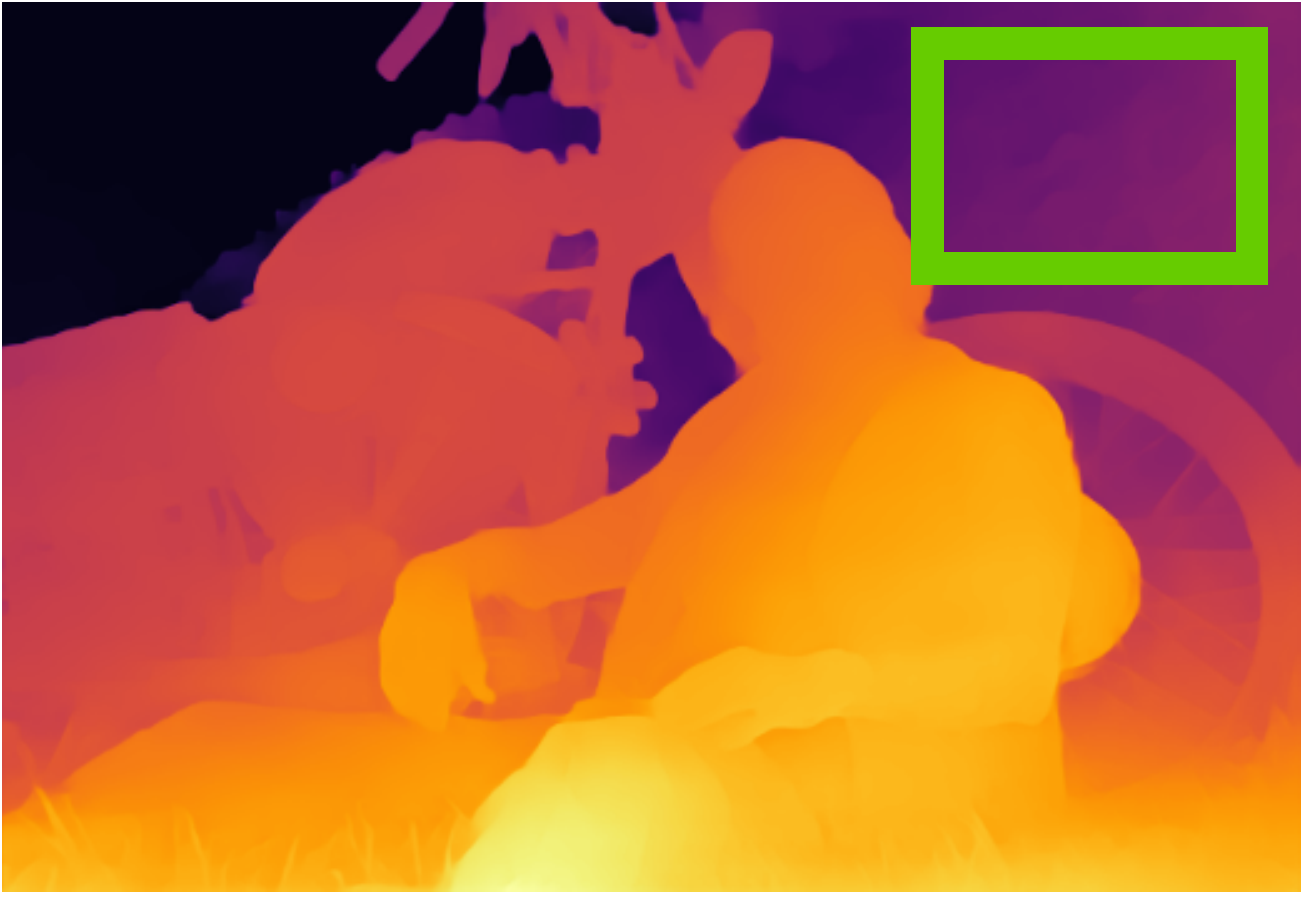}
   \caption{ViT-B + R + AL}
\end{subfigure}
\begin{subfigure}{.16\textwidth}
   \includegraphics[width=0.98\linewidth,height=28.5mm]{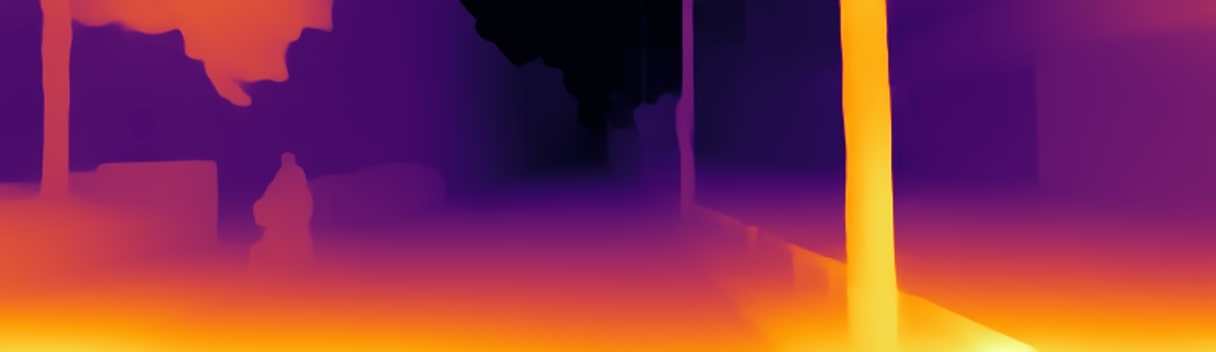} \\ 
   \includegraphics[width=0.98\linewidth,height=28.5mm]{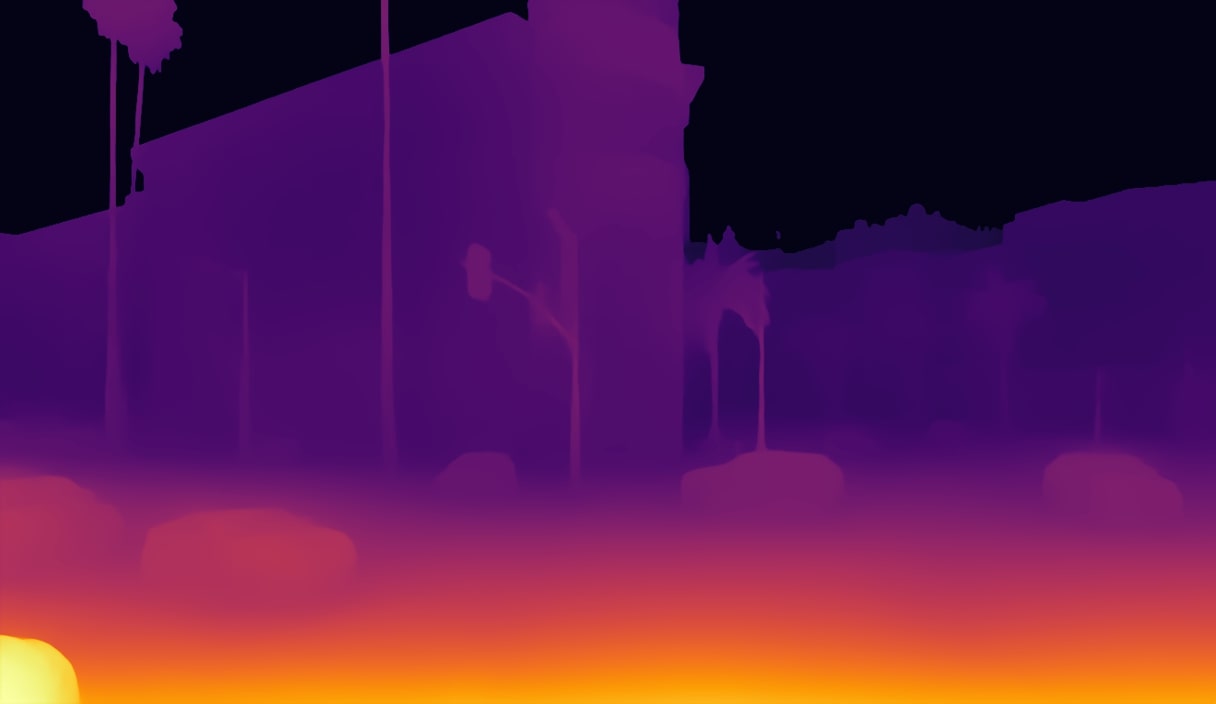}  \\
   \includegraphics[width=\linewidth,height=28.5mm]{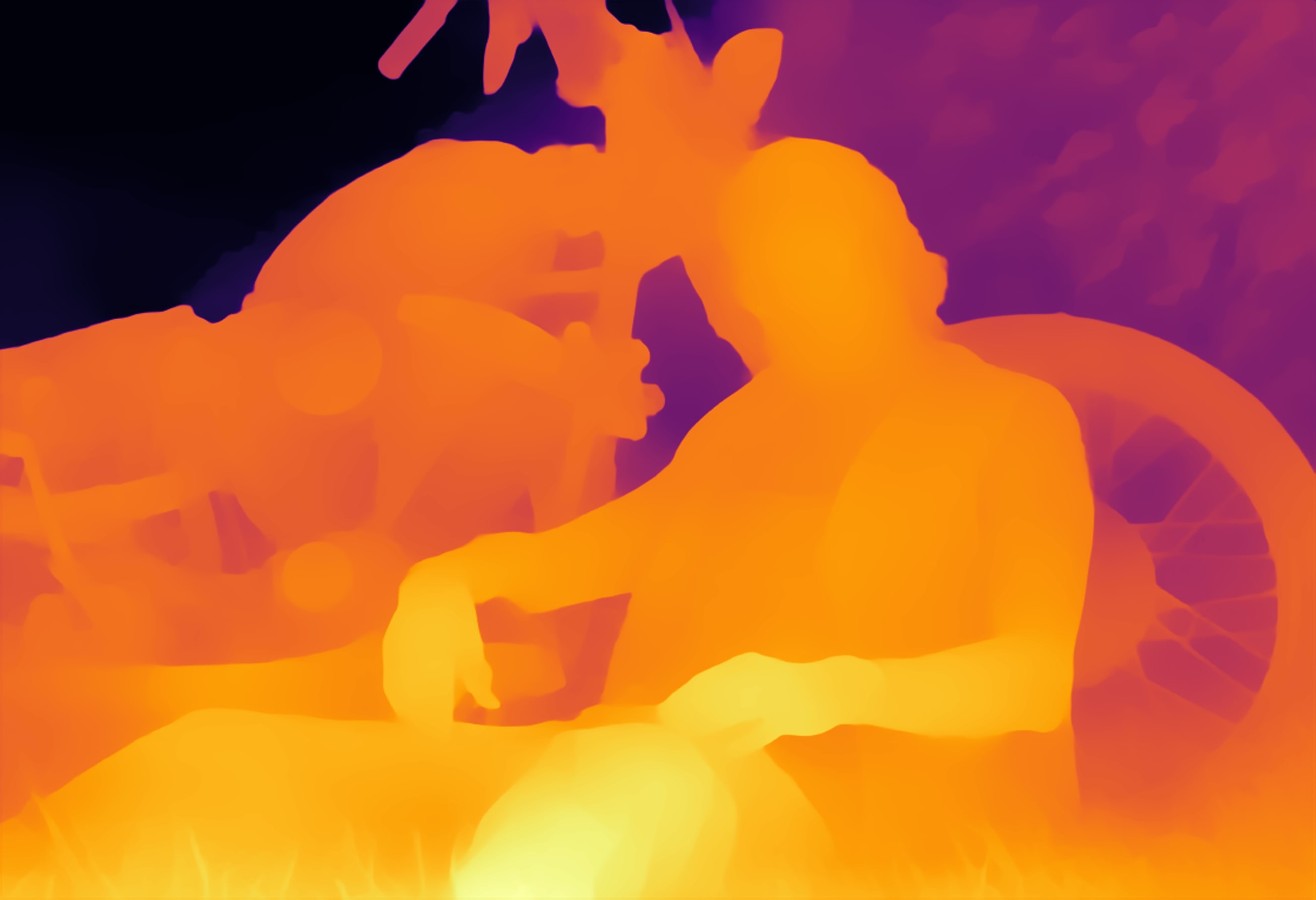}
   \caption{DPT-B + R + AL}
\end{subfigure}
\caption{Outdoor Scenes. $1^{\text {st}}$ Row:- KITTI \cite{Kitti}. $2^{\text {nd}}$Row:- HRSD outdoor. $3^{\text {rd}}$ Row:- RealWorld. Similar to indoor scenes,  our DPT-B + R + AL gives the best performance outputting a consistent depth map with precise overall structure i.e. the motorbike in the real-world image. Original DPT \cite{DPT} again fails to identify objects in the background as shown by the green rectangle i.e. no structure of background buildings in the KITTI image. Multires \cite{multires} leads to the inconsistent depth map, highlighted by green rectangles i.e. depth around the biker's body is fluctuating in the real-world image.}
\label{outdoorimg}
\end{figure*}
\vspace{.2cm}
\textbf{Qualitative Results}
The quantitative results are supported by visual comparisons in Figures \ref{interimg} \& \ref{outdoorimg}. We compare indoor (Figure \ref{interimg}) and outdoor scenes (Figure \ref{outdoorimg}) for analysis and discussion. We also include real-world scenes in both figures to test the performance of various algorithms. In Figures \ref{interimg} \& \ref{outdoorimg}, DPT \cite{DPT} fails to give precise depth edges and lacks details of the structure of background objects. Also, DPT \cite{DPT} fails to get a clear object boundary in real-world scenes. MultiRes \cite{multires} can get a sharper depth map and details but provides inconsistent depth within an object, and these artifacts are highlighted using a rectangular box in Figures \ref{interimg} \& \ref{outdoorimg}. Although the proposed variant ViT-B + R+ AL lacks details in the background, the DPT-B + R + AL gives a more structured and consistent depth with precise object shapes and clearer edges, even for objects further away in the scene. This is reasonable because DPT \cite{DPT} is trained on RGB-D datasets, whereas ViT \cite{ViT} is trained for image recognition tasks. Compared to both DPT \cite{DPT} and MultiRes \cite{multires}, our method results in a smoother depth map closest to ground truth maps.

\par
\par

\vspace{.1cm}
\textbf{Running time analysis}
We further compare the inference time from different algorithms and present them in Table \ref{tb:inferenceSpeed}. It shows the inference speed in milliseconds per frame, averaging over 400 images on the three different resolutions. Timings were conducted on an
Intel I5 $10^{\text {th}}$ generation @ 2. 90GHz with eight physical cores and a single Nvidia RTX A6000 GPU. Multires \cite{multires} take the longest running time as it merges different resolutions to compute a high-quality depth map. %The resolution at which the base estimation is computed, $R_{20}$, and the number of patches merged onto the base estimate is adaptive to the image content. Their method ended up selecting 74. 82 patches per image on average for $R_{20}$ = 1920 × 1080 for the GTA V dataset and 12. 17 patches per image on average for $R_{20}$ = 640*480 for NYU V2. 
DPT \cite{DPT} and the proposed network take similar inference time on smaller resolutions. However, our proposed network is more efficient for high-resolution images due to the fewer patches produced by the feature-extraction module, which is later processed by the transformer layers. 

\begin{table}[h]
\renewcommand{\arraystretch}{1.0} 
	\begin{tabular}{|c|c|c|c|}
	\hline
Method & DPT \cite {DPT} & MultiRes \cite{multires}& Proposed  \\
	\hline\hline
     $640 \times 480$  &17 & 50 & \textbf{14} \\
    $1216 \times 352$  &22 & 80 & \textbf{20} \\
 $1920 \times 1080$ & 55 & 190 & \textbf{38}  \\
		 \hline
	\end{tabular}
	\caption{Inference Speed (ms) per frame. 3 different resolutions are displayed for comparison. Here Proposed implies: DPT-B+R+AL.  }
	\label{tb:inferenceSpeed}
\end{table}

\begin{table}[h] 
\renewcommand{\arraystretch}{1.0}
{\begin{tabular}{|c|c|c|c|}
	\hline
Method  & AbsRel $\downarrow$ &RMSE $\downarrow$ & $\delta$ > 1. 25 $\uparrow$ \\
	\hline\hline
	ViT-32\cite{ViT}&0.137 & 0.494 & 0.803  \\
        ViT-16\cite{ViT}&0.125 & 0.471 & 0.828 \\
        ViT-16+ R-101  &0.112 & 0.387 & 0.852  \\
	ViT-16 + R-50 &\textbf{0.107} & \textbf{0.342} & \textbf{0.882} \\
		 \hline
	\end{tabular}}
	\caption{Feature module analysis. Different variants of ViT \cite{ViT} and Resnet \cite{resnet} are experimented with to choose the best feature extraction module.}
	\label{featurecomp}
\end{table}
\textbf{Feature extraction module analysis}
Here, we compare the effect of the feature extraction module that provides image embedding to our transformer layers, as shown in Figure \ref{fig:overview}. ViT \cite{ViT} uses a simple image flattening technique to transform the image into patches and comes in two variants with ViT-32 and ViT-16. Since we use the Resnet \cite{resnet} feature module to process color RGB images, we compare different pre-trained ResNet encoders, such as Resnet-50 and Resnet-101, with ViT \cite{ViT}. In Table \ref{featurecomp}, we first make a comparison with ViT-16 and ViT-32. We use these weights as an initial encoder weight and train the entire encoder-decoder on the proposed HRSD datasets. We observe that ViT-16 achieves low error and higher accuracy than ViT-32. Later, we fixed ViT-16 as the backbone architecture and experimented with two resnet encoders, such as Resnet-50 and Resnet-101. We again make two different training, one with Resnet-50 with ViT-16 and the other with Resnet-101 as our initial encoder weights, and train the entire encoder-decoder on the proposed HRSD datasets. As shown in Table \ref{featurecomp}, introducing the Resnet encoders significantly improves the performance. However, Resnet-50 outperforms and thus becomes the final choice for our MDE network.

\section{Conclusion}
In this paper, we proposed to generate a
 high-quality synthetic RGB-D datasets from the game GTA-V \cite{gtav} with precise dense depth maps. Since we can control all the aspects of the GTA game, we can capture scenes with varied lighting, different environments, and diverse objects. We trained the DPT \cite{DPT} architecture with DPT \cite{DPT} and ViT\cite{ViT} weights on the proposed HRSD datasets. We observed a significant improvement in the performance of the depth maps, both objectively and subjectively. The performance is further improved by modifying the DPT \cite{DPT} architecture and loss function.   

{\small
\bibliographystyle{ieee_fullname}
\bibliography{egbib}
}
\clearpage
\newpage
\pagebreak

\begin{center}
  \huge{\bfseries --- Supplementary ---}\\
\end{center}
\setcounter{section}{0}
\section{Evaluation Metrics}

Recall for evaluating different depth estimation algorithms; we use Abs-Rel, RMSE, and percentage of correct pixels. Below are the equations used for each error metric:-

\begin{itemize}
\item Absolute relative error (abs-rel):-
\begin{equation}
\frac{1}{N} \sum_{p=1}^{N} \frac{\left|{y}_{p}-\hat{y}_{p}\right|}{\hat{y}_{p}}
\end{equation}
\item Root mean squared error (rmse):-
\begin{equation}
\sqrt{\frac{1}{N} \sum_{p=1}^{N} \frac{\left\|{y}_{p}-\hat{y}_{p}\right\|^{2}}{\hat{y}_{p}}}
\end{equation}
\item Accuracy with threshold $t$: percentage(\%) of $\hat{y}_{p}$, subject to max \begin{equation}
\left(\frac{\hat{y}_{p}}{{y}_{p}}, \frac{{y}_{p}}{\hat{y}_{p}}\right)=\delta<t\left(t \in\left[1. 25\right]\right)
\end{equation}
\end{itemize}
where ${y}_{p}$ and $\hat{y}_{p}$ are the ground-truth depth and the estimated depth at pixel $p$ respectively; N is the total number of pixels of test images. \par

\section{Proposed HRSD datasets}
The proposed HRSD dataset consists of images varying in captured environments and objects (indoor/outdoor scenes, dynamic objects, homogeneous scenes). This enables improved training of monocular depth estimation algorithms leading to a better overall generalization of diverse scenes. We show the statistics of our HRSD dataset in Table \ref{tb:statsgta} which includes details on a number of indoor scenes, outdoor scenes, etc. We generated a total of 100,000 pairs of RGB and ground truth depth. 
\begin{table}[h]
\renewcommand{\arraystretch}{1.0} 
	\begin{tabular}{|c|c|c|c|}
	\hline
HRSD (1920*1080) & Train & Validation & Test  \\
	\hline\hline
     Indoor  &29,000 & 5000 & 4000 \\
    Outdoor  & 46000 & 10000 & 6000 \\
     \hline
	\end{tabular}
	\caption{Statistics of Proposed HRSD datasets}

	\label{tb:statsgta}
\end{table}

\par
\textbf{Scripthook V} Recall we used G2D \cite{g2d} to extract depth maps from Gbuffers during game-play in the rendering pipeline. We also want to credit ScriptHook V as G2D is formulated on Scripthook V \cite{scripthook}, an Alexander Blade library that allows access to GTA-V's native functionality. Adopting Scripthook V distinguishes our dataset from Richter et al. \cite{gtasemand} to create modifications ("mods") for meddling within the detailed virtual environment of GTA-V. G2D allows users to collect hyper-realistic computer-generated imagery of an urban scene under controlled 6DOF camera poses and varying environmental conditions. Users directly interact with G2D while playing the game; specifically, users can manipulate the conditions of the virtual environment on the fly. The original aim of Scripthook V is to provide a framework to construct modifications (“mods") to the game. Currently, a wide range of fascinating mods is available,  e.g., the Invisibility Cloak, which can make the protagonist invisible. The list of native functions supported by Scripthook V can be found on \cite{mods}.

\section{Ablation Studies}
Recall that we propose to use an attention-based supervision loss (AL) term in addition to the default $L_{1}$ loss in the proposed algorithm. Here, we discuss the effects of choosing different loss functions for training DPT \cite{DPT} networks on the proposed HRSD datasets. 

\par 
\textbf{Effect of loss  function}
 To verify the effectiveness of the additional attention loss term, we trained the model (DPT-B + R) with three different loss terms, and they are as follows: 
 
 \emph{1)} L1 loss term, \emph{2)} DPT's \cite{DPT} loss and \emph{3)} $L_{1}$+ $L_{AL}$. 
 
This study is performed on a reduced HRSD dataset of only 30000 images. We used 25000 images for training and 5000 images for the validation set. 
We report results on validation sets with the metric described in equation (1) - (3), and the results are shown in Table \ref{tb:quant}. 

From Table \ref{tb:quant}, it is obvious that the $L_{1}$+ $L_{AL}$ gives a relatively low error and higher accuracy compared to other loss functions, demonstrating attention-loss effectiveness.

\begin{table}[h] 
	\centering
	\begin{tabular}{|c|c|c|c|}
	\hline
Method  & AbsRel $\downarrow$ &RMSE $\downarrow$ & $\delta$ > 1. 25 $\uparrow$ \\
        \hline\hline
     $L_{DPT}$ \cite{DPT} &0.107 & 0.394 & 0.872 \\ 
	$L_{1}$&0.095 & 0.324 & 0.891 \\
   $L_{1}$+ $L_{AL}$&\textbf{0.074} & \textbf{0.288} & \textbf{0.921}\\ 
		 \hline
   	\end{tabular}
	\caption{Evaluation of different loss functions on the proposed HRSD datasets. }
	\label{tb:quant}
\end{table}

\end{document}